\PassOptionsToPackage{table}{xcolor}
\PassOptionsToPackage{capitalize}{cleveref}
\PassOptionsToPackage{hyphens}{url}
\documentclass[11pt]{jarvisstyle}
\setcitestyle{numbers,square,comma}

\usepackage[utf8]{inputenc}
\usepackage{url}
\usepackage{amsmath,amssymb,amsfonts}
\usepackage{nicefrac}
\usepackage{fontawesome5}
\usepackage{capt-of,afterpage}
\usepackage{enumitem,array,siunitx,tabularx,colortbl,pifont,makecell}
\usepackage{xspace,adjustbox,threeparttable,algorithm,algpseudocode,wrapfig}
\usepackage[normalem]{ulem}
\makeatletter
\def\paragraph{\@startsection{paragraph}{4}{\z@}{0.6ex plus 0.3ex minus .2ex}{-1em}{\normalsize\bfseries}}
\makeatother
\renewcommand\authorformat[2][]{{\bfseries\fontsize{9}{10.4}\selectfont #2$^{#1}$}}
\renewcommand\affiliationformat[2][]{{\fontsize{8}{10}\selectfont $^{#1}$#2}}
\newenvironment{ack}{\section*{Acknowledgments}}{}

\newcommand{\answerTODO}[1][]{\textcolor{red}{\bfseries [TODO]}}
\newcommand{\justificationTODO}[1][]{\textcolor{red}{\bfseries [TODO]}}

\definecolor{lightpurple}{RGB}{245, 242, 255}
\newcommand{\greencheck}{\textcolor{green}{\ding{51}}}

\newcommand{\redcross}{\textcolor{red}{\ding{55}}}
\newcommand{\halfcheck}{\rlap{\textcolor{orange}{\ding{51}}} \hspace{-0.38em} {\textcolor{orange}{\ding{55}}}}

\definecolor{okgreen}{HTML}{1B8A3A}
\definecolor{nored}{HTML}{C0392B}
\newcommand{\yes}{\textcolor{okgreen}{\ding{51}}}
\newcommand{\no}{\textcolor{nored}{\ding{55}}}

\definecolor{bandExt}{HTML}{E8EEF7}
\definecolor{bandInt}{HTML}{F4ECE4}
\definecolor{bandQual}{HTML}{ECF2E8}

\definecolor{tabHeader}{HTML}{1F3A68}
\definecolor{tabHeaderText}{HTML}{FFFFFF}
\definecolor{rowAlt}{HTML}{F6F8FB}
\definecolor{bandGemini}{HTML}{E6EFF7}
\definecolor{bandGPT}{HTML}{F0EAE0}
\definecolor{bandQwen}{HTML}{E9F1E5}
\definecolor{bandGemma}{HTML}{F4ECF2}
\definecolor{bandClaude}{HTML}{FBEFE0}
\definecolor{bandHigh}{HTML}{D6E4D2}

\newcolumntype{Y}{>{\raggedright\arraybackslash\hspace{0pt}}X}

\newcommand{\hdrcell}[1]{\textcolor{tabHeaderText}{\textbf{#1}}}

\definecolor{famPerf}{HTML}{1F3A68}
\definecolor{famEff}{HTML}{8B5A1C}
\definecolor{famQual}{HTML}{3D5A2E}

\title{\mbox{%
  \raisebox{-0.30em}[0pt][0pt]{\includegraphics[trim=74bp 68bp 74bp 58bp,clip,height=1.30em]{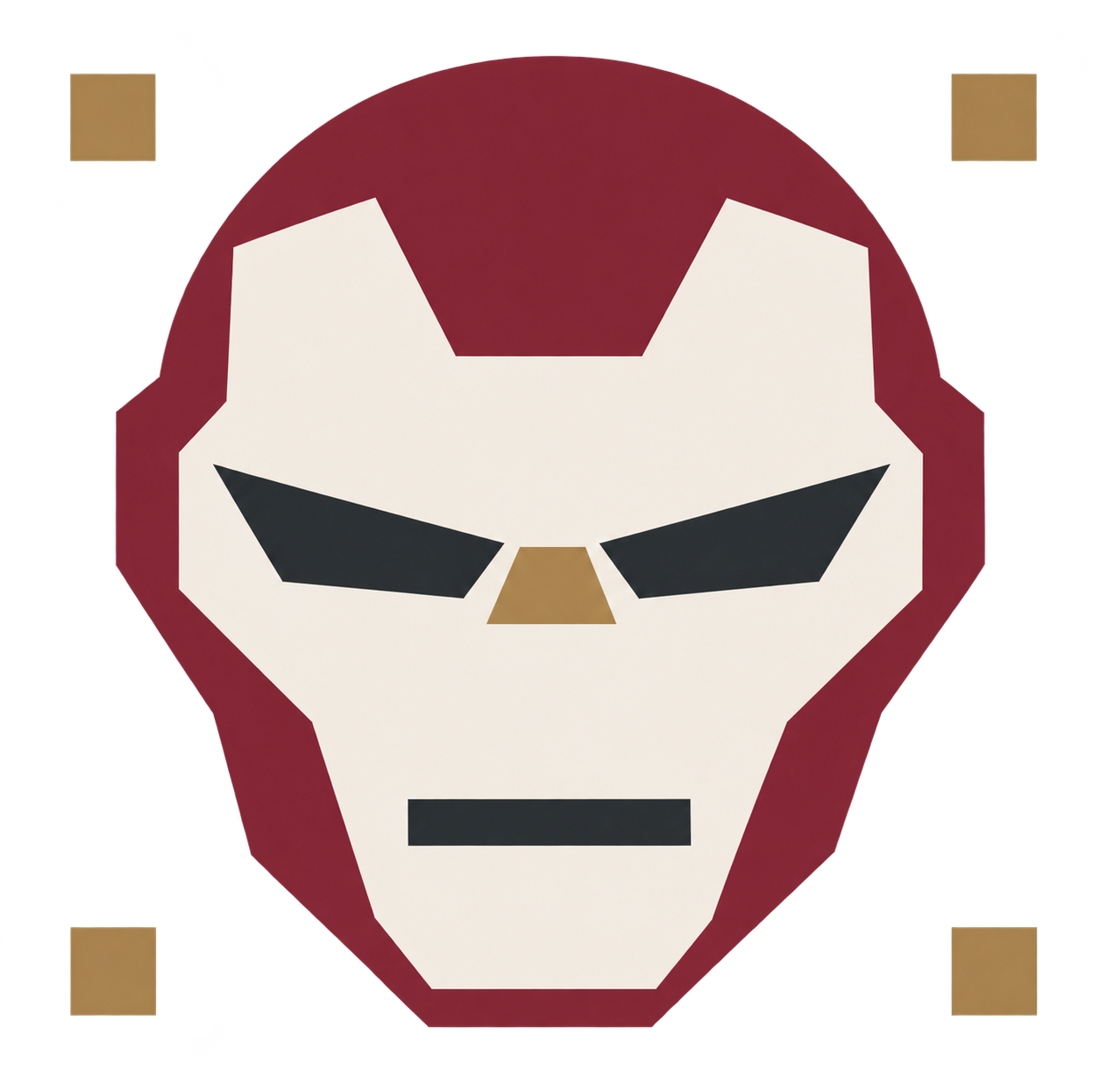}}%
  \hspace{0.35em}JARVIS-Bench}\\[-2pt]
  {\fontsize{12}{13}\selectfont Benchmarking Personal Intelligence Agents on Long-Horizon Real-User Daily Traces}}
\author[1,*,\dagger]{Weizhi~Zhang}
\author[1,*,\dagger]{Wei-Chieh~Huang}
\author[2,*]{Yueqing~Liang}
\author[3,4]{Liwei~Jiang}
\author[5]{Zhengxiang~Wang}
\author[6]{Liangwei~Yang}
\author[7,8]{Zechen~Li}
\author[4]{Yuchen~Wu}
\author[9]{Haozhen~Zhang}
\author[10]{Yu~Wang}
\author[11]{Yuanchen~Bei}
\author[12]{Yue~Zhou}
\author[11]{Siqi~Zhu}
\author[1]{Henry~Peng~Zou}
\author[13]{Jiahong~Liu}
\author[13]{Xinni~Zhang}
\author[14]{Paul~Martin}
\author[15,16]{Joseph~Marvin~Imperial}
\author[17]{Yuyang~Luo}
\author[2]{Xiongxiao~Xu}
\author[17]{Baixiang~Huang}
\author[17]{Shanglin~Wu}
\author[14]{Lucas~Resck}
\author[19]{Yibo~Wang}
\author[1]{Yuqing~Liu}
\author[1]{Langzhou~He}
\author[1]{Chengze~Li}
\author[18]{Yuxin~Tian}
\author[1]{Kening~Zheng}
\author[1]{Enze~Ma}
\author[20]{Boi~Huynh}
\author[14]{Zheng~Hui}
\author[11]{Rui~Yang}
\author[11]{Tao~Feng}
\author[1]{Jie~Yang}
\author[1]{Shanghao~Li}
\author[17]{Haoran~Wang}
\author[1]{Wooseong~Yang}
\author[7,1]{Hanrong~Zhang}
\author[1]{Huanhuan~Ma}
\author[1]{Daye~Yoon}
\author[21]{Yaozu~Wu}
\author[22]{Shikan~Lian}
\author[1]{Xiaoqian~Ruan}
\author[1]{Fangxin~Wang}
\author[1]{Hyeonjeong~Park}
\author[23,18]{Hins~Hu}
\author[1]{Wenzhe~Fan}
\author[24]{Chen~Wang}
\author[25]{Yifan~Wang}
\author[26]{Jintao~Huang}
\author[27]{Yijun~Wang}
\author[28]{Haixin~Wang}
\author[29]{Yating~Wu} \author[11]{Cheng~Qian}
\author[19]{Yifan~Gu}
\author[21,30]{Dongyuan~Li}
\author[31]{Song~Wang}
\author[4]{Aylin~Caliskan}
\author[32]{Jindong~Wang}
\author[33]{Xiao~Luo}
\author[34]{Yankai~Chen}
\author[17]{Kai~Shu}
\author[34,35]{Xue~Liu}
\author[36,37]{Yuexing~Hao}
\author[27,37]{Xiaomin~Li}
\author[1,\dagger]{Philip~S.~Yu}

\affiliation[1]{University of Illinois Chicago}
\affiliation[2]{Illinois Institute of Technology}
\affiliation[3]{NVIDIA}
\affiliation[4]{University of Washington}
\affiliation[5]{Stony Brook University}
\affiliation[6]{Salesforce AI Research}
\affiliation[7]{Google Research}
\affiliation[8]{University of New South Wales}
\affiliation[9]{Nanyang Technological University}
\affiliation[10]{University of California, San Diego}
\affiliation[11]{University of Illinois Urbana-Champaign}
\affiliation[12]{Capital One}
\affiliation[13]{The Chinese University of Hong Kong}
\affiliation[14]{University of Cambridge}
\affiliation[15]{University of Bath}
\affiliation[16]{Pivotal Research}
\affiliation[17]{Emory University}
\affiliation[18]{Meta}
\affiliation[19]{Amazon}
\affiliation[20]{University of Information Technology, Vietnam National University}
\affiliation[21]{The University of Tokyo}
\affiliation[22]{Independent Researcher}
\affiliation[23]{Cornell University}
\affiliation[24]{Roblox}
\affiliation[25]{Purdue University}
\affiliation[26]{The Ohio State University}
\affiliation[27]{Harvard University}
\affiliation[28]{University of California, Los Angeles}
\affiliation[29]{The University of Texas at Austin}
\affiliation[30]{Institute of Science Tokyo}
\affiliation[31]{University of Central Florida}
\affiliation[32]{William \& Mary}
\affiliation[33]{University of Wisconsin--Madison}
\affiliation[34]{MBZUAI}
\affiliation[35]{McGill University}
\affiliation[36]{Massachusetts Institute of Technology}
\affiliation[37]{MatrAIx}

\contribution[*]{Equal contribution}
\contribution[\dagger]{Corresponding authors}
\abstract{
Personal intelligence agents aim to support users across daily-life tasks by learning their preferences, routines, needs, and emotional states from long-term interaction.
They are expected to anticipate user needs, surface help proactively, and infer preferences a user has never explicitly articulated. Existing benchmarks for LLM personalization lack real user data spanning extended periods across various domains in real-world agentic settings, and typically reduce personalization to simplified tasks such as recommendation and fact retrieval.  We introduce \textsc{Jarvis-Bench}, a new  benchmark for evaluating personal intelligence agents, curated from real users' daily traces and augmented with synthetic trajectory expansions grounded in implicit-preference insights from surveys and structured personas. Specifically, 
\textsc{Jarvis-Bench} is built from participants who contributed 28 to 44 days of multimodal self-tracking logs, structured personas, and a 100-question implicit-preference survey spanning eight domains. 
\textsc{Jarvis-Bench} supports four downstream task surfaces: Implicit Preference Reasoning, Proactive Help Prediction, Emotion-Aware Modeling, and Cross-Domain Transfer, all consumed through a single uniform event iterator and evaluation pipeline with long-horizon user memory. We benchmark 54 LLMs from 11 model families (e.g., GPT, Claude, Gemini, Qwen, Grok) across a five-condition input ladder, ranging from no memory to lifelong memory.
Results show that memory and model scaling improve personalization, yet current frontier models remain far from robust personal intelligence within the population covered by our dataset.
}
\renewcommand{\metadatalist}{{\centering\small\bfseries
  \href{https://agentmemoryworld.github.io/jarvis-bench/}{\faGlobe\hspace{0.35em}Website}\hspace{2em}
  \href{https://github.com/AgentMemoryWorld/JARVIS-Bench}{\faGithub\hspace{0.35em}Code}\hspace{2em}
  \href{https://huggingface.co/datasets/WZDavid/JARVIS-Bench}{\raisebox{-0.18em}{\includegraphics[height=1.15em]{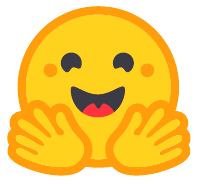}}\hspace{0.35em}Data}\par}}
\begin{document}
\maketitle

\noindent\begin{minipage}{\textwidth}
  \centering
  \includegraphics[width=\textwidth]{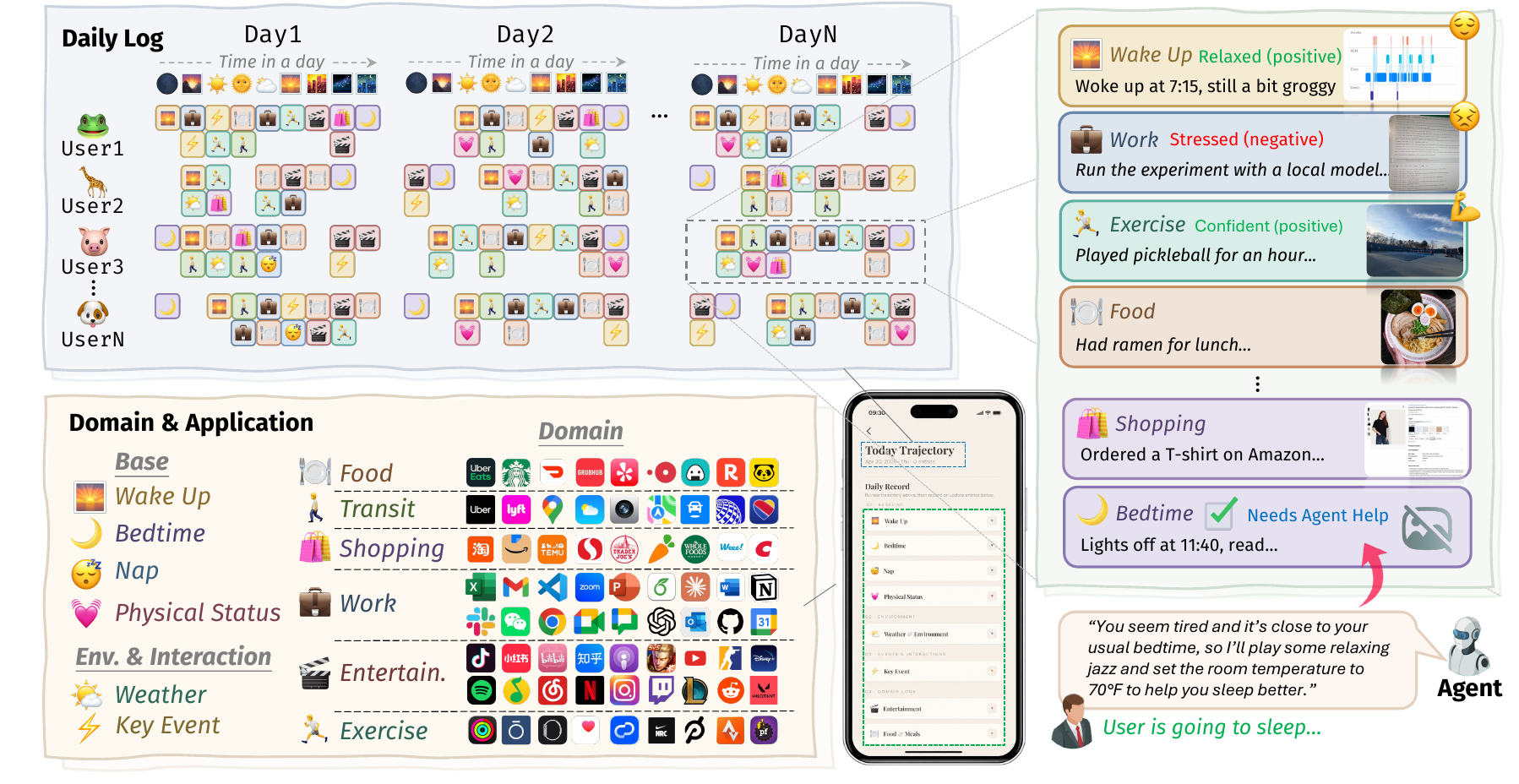}
  \captionsetup{font=small,skip=4pt}
  \captionof{figure}{Overview of \textsc{Jarvis-Bench} regarding real-user traces and ground-truth feedback collection process. The benchmark collects long-horizon daily logs through a mobile web app, organizing multimodal life events across time, users, and domains such as food, work, exercise, entertainment, shopping, and bedtime. These structured traces provide implicit preference, emotion, and proactive-help signals across domains for evaluating personal intelligence agents in realistic daily-life settings over extended periods of interaction.}
  \label{fig:main_figure}
\end{minipage}

\clearpage
\begingroup
\renewcommand{\thefootnote}{}
\makeatletter
\renewcommand{\@makefnmark}{} \long\def\@makefntext#1{\parindent=0pt\noindent#1}
\makeatother
\interfootnotelinepenalty=10000
\footnotetext{%
  \fontsize{8}{9.5}\selectfont
  \setlength{\parskip}{2pt}
  \noindent\textbf{Author affiliations}\par\nobreak
  \affiliationlist
}
\endgroup

\section{Introduction}
\label{sec:intro}

As AI agents~\cite{luo2025large} become increasingly integrated into everyday applications, their role is shifting from task-oriented tools~\cite{jimenez2023swe,zhou2023webarena} to more personalized and proactive assistants. 
This vision is perhaps best captured by J.A.R.V.I.S., Tony Stark's AI assistant in the \textit{Iron Man} franchise, capable of recording, understanding, anticipating, and acting for the user.
Recent agents have begun to incorporate user profiles~\cite{salemi2024lamp,zhang2026personaagent} or domain-specific interaction histories~\cite{tan2025personabench,zhang2026memorycd}, but such personalization remains limited: personal intelligence requires the more fundamental ability to understand users' implicit preferences.
Such agents must learn from real behavior over time, infer preferences and routines, anticipate emotional needs, and adapt across contexts and scenarios. 
This form of personalization is inherently implicit, long-horizon, and cross-domain: it emerges from distributed behavioral signals rather than explicit user input, and is central to transforming agents from reactive tools into proactive personal companions~\cite{lu2024proactiveagent,yang2025contextagent} that support users across weeks of their daily lives.

However, existing personalization benchmarks are insufficient for evaluating personal intelligence agents (more detailed comparisons are in Table~\ref{tab:benchmark_comparison_updated} in Appendix~\ref{app:related-full}). 
On the one hand, most benchmarks focus on narrow, task-specific settings, such as recommendation or dialogue, and do not capture whether an agent can truly understand a user across diverse real-world contexts~\cite{zhang2018personalizing,maharana2024evaluating,salemi2024lamp,tan2025personabench}. 
On the other hand, many personalization benchmarks rely heavily on fully simulated data, despite growing evidence that synthetic user behavior diverges significantly from real-world heterogeneity~\cite{hu2025simbench,zhou2026sim2real,peng2026funhouse}.
More fundamentally, these benchmarks often miss the core challenge of personalization in practice: user preferences are not merely \emph{\textbf{explicit facts}} to be retrieved from long context, but \emph{\textbf{latent patterns implicitly expressed}} through decisions, routines, emotions, and interactions distributed across time and domains.
As a result, existing benchmarks do not adequately measure whether models can infer and act on such latent preferences from real-world behavior, leaving a critical gap in evaluating personal intelligence agents~\cite{zhao2025llms,tsaknakis2025llms}.
To address this gap, we move beyond task-specific and simulated settings, and construct a benchmark grounded in real users' daily-life behavior to enable evaluation of personalization in complex, real-world scenarios.

We introduce \textsc{{Jarvis-Bench}}, a benchmark for evaluating personal intelligence agents grounded in longitudinal real-user behavior. 
Unlike prior work that relies on simulated users or explicit preference annotations, both long-context evidence input and ground-truth labels in \textsc{Jarvis-Bench} are derived directly from real users.
As shown in Figure~\ref{fig:main_figure}, we deploy an IRB-approved custom mobile interface to 70 consenting participants, of whom 42 contribute 28 to 44 days of multimodal daily-life trajectories under a fully anonymous data-collection protocol. 
Each user provides rich supervision signals, including behavioral logs across diverse domains, structured personas, implicit-preference surveys, emotion annotations, and proactive-help intention requests. 
To overcome the sparsity of real-world data while preserving fidelity, we design a synthesis pipeline that expands behavioral trajectories without explicitly restating preferences through a four-stage process, ensuring that preferences are expressed only through observable actions and contextual signals. 

\paragraph{What \textsc{Jarvis-Bench} evaluates and measures.}
Using this benchmark, we evaluate personal intelligence through a unified framework that captures four core capabilities: Implicit Preference Reasoning, Proactive Help Prediction, Emotion-Aware Modeling, and Cross-Domain Transfer. All tasks are grounded in the same behavioral trajectories, enabling systematic analysis of how models utilize user context under varying memory conditions~\cite{huang2026rethinking,huang2026harness}. Our evaluation across 54 large language models (LLMs) reveals three key findings, under the assumption that our qualified real-user cohort provides a meaningful first approximation of broader personalization patterns through longitudinal daily traces and user-provided supervision. First, behavioral trajectories contribute substantially more than explicit persona descriptions, indicating that preference inference is primarily grounded in observed user behavior. Second, even with oracle-level memory, the best-performing model achieves only 73.2\% accuracy, suggesting that inferring latent preferences from behavior remains far from solved. Third, no model consistently performs well across all capabilities, highlighting the inherently multi-dimensional nature of personal intelligence. Together, these results demonstrate that current LLMs, despite strong general capabilities, still struggle to understand the real users in our cohort, establishing \textsc{Jarvis-Bench} as a principled framework for evaluating this frontier. In summary, we make the following three contributions toward evaluating personal agents:

\begin{itemize}[leftmargin=*]
\item We formalize personal intelligence as a unified ability to reason and act on user preferences from long-horizon behavioral traces, and identify four core capabilities, including Implicit Preference Reasoning, Proactive Help Prediction, Emotion-Aware Modeling, and Cross-Domain Transfer.
\item We introduce \textsc{Jarvis-Bench}, a benchmark grounded in real-user daily behavior, and design a full-stack evaluation pipeline, including IRB-approved privacy-preserving data collection, standardized clue-planted trajectory synthesis, a memory-condition ladder, and four task settings.
\item Through a large-scale evaluation of 54 LLMs from 11 model families, we show that scaling user memory and model capacity improves personalization but remains far from sufficient: the best model reaches only 55.0\% accuracy with the long trajectory, proactive and emotion-aware reasoning remain weak, and cross-domain transfer is uneven and sometimes noisy.
\end{itemize}

\begin{figure}[!t]
  \centering
  \includegraphics[width=\textwidth]{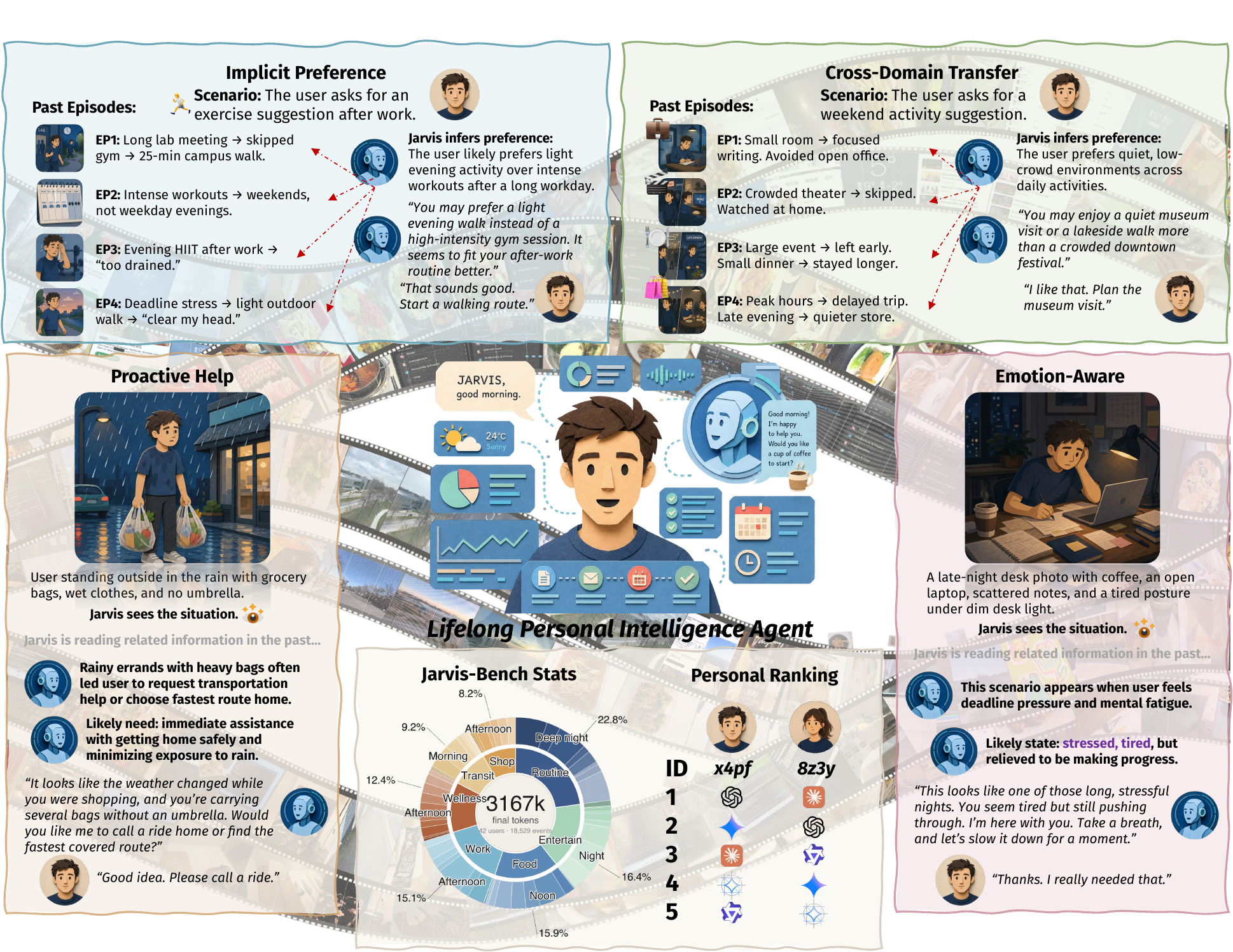}
  \caption{Illustration of \textsc{Jarvis-Bench}'s four task surfaces for evaluating personal intelligence agents: Implicit Preference Reasoning, Proactive Help Prediction, Emotion-Aware Modeling, and Cross-Domain Transfer. 
More statistical details are attached in Figure~\ref{fig:composition-tokens}.
}
  \label{fig:overview}

\end{figure}

\section{Dataset Overview and Task Formulation}
\label{sec:overview}
Personalization signal about a real individual naturally comes from two complementary sources: the fine-grained daily behaviours that show how a person actually lives, and the distilled survey and persona summaries that capture their stable preferences. \textsc{Jarvis-Bench} brings the two together into a single longitudinal record that keeps the granularity of real behaviour while letting distilled preferences surface only as behavioural consequences, and evaluates that record through a unified event iterator over four task surfaces and a five-condition input ladder.

\subsection{Released artefact and dataset construction}
\label{sec:overview-schema}

The released artefact for each of the 42 retained participants is a longitudinal record of day-indexed events under a single typed schema. \emph{Anchor events} reproduce the participant's original log entries verbatim and provide the ground truth for the proactive-help and emotion tasks; \emph{synthetic events} densify the trajectory where self-tracking is naturally sparse, conditioned on day theme, narrative continuity, the participant's behavioural signature, and a planting plan that designates which events express which preference. Each participant contributes three forms of supervision: multi-day daily logs over 12 event subdomains, a structured persona with a free-text self-description, and a 100-question implicit-preference instrument over 8 lifestyle domains (\cref{tab:domain-mapping} gives the canonical mapping between the two levels). A four-stage clue-planting synthesis pipeline assembles the trajectory under two contracts: anchor content is preserved verbatim, and survey responses enter only through a distilled insight object so that synthetic events encode preferences solely as behavioural consequences. Each released trajectory passes a three-stage quality bar whose final stage is a participant-driven fidelity confirmation on a randomly sampled subset of synthesised events. The released set covers approximately 440 events per participant (about half anchor and half synthetic) over 28 to 44 days; full schema, pipeline, validator, taxonomy, and feedback-loop details are in \cref{app:dataset-construction}.

\subsection{Task formulation and input-condition ladder}
\label{sec:overview-tasks}

\textsc{Jarvis-Bench} supports four task surfaces, summarised in \cref{tab:task-table}. All four tasks share the same event iterator and differ only in (i) which subset of events is the prediction target and (ii) which ground-truth label is consumed: \textbf{Implicit Preference Reasoning} tests \emph{declarative} preference recovery from behavioural evidence, \textbf{Proactive Help Prediction} tests \emph{timing-conditioned trigger} judgement against the participant's own \texttt{agentHelp} ground truth, \textbf{Emotion-Aware Modeling} tests \emph{scene-level affect} reasoning against participant-reported mood annotations, and \textbf{Cross-Domain Transfer} tests whether preference signal in one lifestyle domain transfers to others. The four surfaces operationalise four decisions that a longitudinal personal agent must make: what the user prefers, when assistance is warranted, how the user feels, and whether a preference carries over to a new domain. They admit different failure modes, since an agent can infer a preference correctly yet intervene at the wrong moment or ignore the user's current mood, so no surface is a proxy for another.

\definecolor{taskA}{HTML}{F0F6FB}   
\definecolor{taskB}{HTML}{F2F7F0}   
\definecolor{taskC}{HTML}{FAF2F2}   
\definecolor{taskD}{HTML}{F4EFF6}   
\begin{table}[t]
\centering
\caption{\textbf{Task Card.} All four tasks consume the same trajectory artefact through unified event iterators. Implicit Preference Reasoning uses the full five-level memory ladder (\cref{tab:conditions}); Proactive Help Prediction and Emotion-Aware Modeling use the user history preceding the target; Cross-Domain Transfer uses the S1 / S2 ablations derived from C2 and C0 (\cref{sec:results-transfer}).}
\label{tab:task-table}
\small
\setlength{\tabcolsep}{4pt}
\renewcommand{\arraystretch}{1.20}
\arrayrulecolor{tabHeader}
\begin{tabularx}{\linewidth}{@{}>{\bfseries}p{1.8cm}
  >{\columncolor{taskA}}Y
  >{\columncolor{taskB}}Y
  >{\columncolor{taskC}}Y
  >{\columncolor{taskD}}Y@{}}
\toprule
\rowcolor{tabHeader}
\hdrcell{}
  & \hdrcell{$\star$~Implicit Preference}
  & \hdrcell{$\triangleright$~Proactive Help}
  & \hdrcell{$\heartsuit$~Emotion-Aware}
  & \hdrcell{\mbox{$\leftrightarrow$~Domain Transfer}} \\
\midrule
Input
  & user memory under the full ladder C0--C4 (no memory, short or long trajectory, with or without persona)
  & all user history before the anchor's day (in full detail) plus the anchor event, optionally with its photograph
  & all prior days in full plus earlier same-day events, then the target event, optionally with its photograph
  & long trajectory with one lifestyle domain removed (S1) or only one lifestyle domain retained (S2) \\
Ground truth
  & 100 survey answers across 8 lifestyle domains
  & \texttt{agentHelp} $\in \{0,1\}$ and \texttt{agentHelpReason}
  & \texttt{moodKeywords} and the polarity derived from them
  & 100 survey answers across 8 lifestyle domains \\
Metric
  & accuracy, reported overall and stratified by lifestyle domain, clue difficulty, clue type, and planted vs.\ unplanted evidence
  & F1 and rationale match rate (RMR), each dual-reported as per-user macro-average and pooled-anchor micro-average
  & polarity accuracy and mood-F1, each dual-reported as per-user macro-average and pooled-event micro-average
  & diagonal $\Delta$-accuracy relative to C2 (S1 drop-one) or relative to C0 (S2 keep-only) \\
\bottomrule
\end{tabularx}
\arrayrulecolor{black}
\end{table}

\label{sec:overview-ladder}
To isolate \emph{which} kind of memory drives personalization gains, Implicit Preference Reasoning uses a five-level ladder of cumulative user memory: \textbf{No Memory} (C0, no user context), \textbf{Short Trajectory} (C1, first 14 days of behavioural log), \textbf{Long Trajectory} (C2, full trajectory without persona), \textbf{Persona Memory} (C3, full trajectory plus the Stage-1 persona profile), and \textbf{Lifelong Memory} (C4, full trajectory plus persona plus a distilled survey-insight oracle ceiling); per-condition contents are in \cref{tab:conditions} (\cref{app:ipq}). The survey insight is a condensed and deliberately vague summary that names no survey question or option, so even C4 requires combining it with behavioural evidence rather than reading answers back. Removing the synthetic clues lowers C4 accuracy by 3 to 6 points (\cref{sec:app-val-anchor}). The Cross-Domain Transfer task additionally uses S1 / S2 ablations derived from C2 and C0 (\cref{sec:results-transfer}), which remove or keep a single lifestyle domain.

\section{Benchmark Tasks and Experimental Settings}
\label{sec:experiment-design}
We evaluate the four task surfaces of \textsc{Jarvis-Bench} (cf.\ \cref{tab:task-table}) with a unified five-level user-memory ladder, comprising No Memory (C0), Short Trajectory (C1), Long Trajectory (C2), Persona Memory (C3), and Lifelong Memory (C4); the per-condition contents are tabulated in \cref{tab:conditions}. Implicit Preference Reasoning runs at all five levels, Proactive Help and Emotion-Aware Modeling use the history preceding the target, and Cross-Domain Transfer additionally introduces two ablation conditions, S1 and S2, derived from C2 and C0 respectively. The remainder of this section details the evaluation protocol of each task and the experimental setup.

\textbf{Implicit Preference Reasoning (evaluation protocol).}
For each participant, the model predicts a single answer choice in $\{A, B, C, D\}$ for every survey question, with abstention scored as zero. Per-user accuracy is computed against the participant's ground-truth answer key, and the macro-average is taken across the 42 retained users. We additionally report stratified accuracies by lifestyle domain (eight categories, see \cref{tab:domain-mapping}), \texttt{clue\_difficulty} (planted clues are restricted to questions labelled \texttt{hard} or \texttt{very\_hard}), \texttt{clue\_type} (three main types), and planted versus unplanted evidence; full breakdowns are reported in \cref{app:ipq}.

\textbf{Proactive Help Prediction (evaluation protocol).}
The binary classification pool holds the image-bearing anchors of every day that has an anchor with a photograph and a non-empty \texttt{agentHelpReason}. Conditioned on the user's full trajectory over all prior days, the model sees that day's candidate anchors, either as facts-only text or as their photographs, and is asked to predict for each whether the user would benefit from proactive assistance at that moment, accompanied by a free-text rationale. The headline metrics are F1 against the participant's binary flag and the rationale match rate (RMR), where an LLM judge scores the semantic alignment between the model's rationale and the participant's \texttt{agentHelpReason}. The judge agrees with the majority vote of five human annotators at Cohen's $\kappa = 0.91$ (\cref{sec:app-val-judge}). Because user coverage on this task is heterogeneous, we report two complementary aggregates: a per-user macro-average over the 15 covered users and a pooled micro-average over the 254 anchors with $n_{\text{pos}}=111$. A user is covered if at least one of their logged days contains such an anchor. Fifteen of the 42 users qualify, and their Implicit Preference accuracy does not differ significantly from that of the other 27 (\cref{sec:app-val-selection}). The two aggregates can diverge substantially when anchor counts are uneven across users, motivating the dual-reporting protocol used throughout this section.

\textbf{Emotion-Aware Modeling (evaluation protocol).}
The emotion pool holds the 123 anchor events (78 positive, 22 negative, 23 mixed) in the covered users' last five logged days that carry both a photograph and non-empty \texttt{moodKeywords}, drawn from 15 covered users selected as in \cref{sec:app-val-selection}. Conditioned on the user's prior-day trajectory together with the same-day events that precede the target event, the model is presented with the target event as facts-only text or as its photograph alone, and is asked to predict a single polarity label and a multi-label mood vector. The headline metrics are polarity accuracy (three-class, exact match against the inferred polarity) and mood-F1 over the multi-label mood vocabulary. Per-user macro-averages and pooled micro-averages are reported jointly under the same protocol as the proactive task.

\textbf{Cross-Domain Transfer (evaluation protocol).}
Two ablation conditions are derived from the Long Trajectory (C2) and No Memory (C0) baselines. The drop-one condition (S1) removes from the Long Trajectory every event whose subdomain maps to a target lifestyle domain $X$ (per \cref{tab:domain-mapping}); the model is then asked the full 100 survey questions, and the diagonal cell, in which the dropped and asked domains coincide, quantifies the \emph{necessity} of $X$. The keep-only condition (S2) starts from No Memory and adds back only those events whose subdomain maps to $X$; the diagonal cell quantifies the \emph{self-sufficiency} of $X$. Residual events whose subdomain does not map to any of the seven lifestyle domains are retained in both ablations to preserve narrative continuity. The reported metric is the diagonal $\Delta$-accuracy relative to the corresponding baseline (\cref{sec:results-transfer}).

\subsection{Models and conditions}
\label{sec:exp-setup}

We evaluate 54 LLMs from 11 model families on Implicit Preference Reasoning: Claude (Haiku-4.5 and 5.5, Sonnet-4.6, 5, and 5.5, Opus-4.5 to 4.8, 5, and 5.5, Fable-5 and 5.1), GPT (GPT-5 Nano, Mini, and full, GPT-5.4, GPT-5.5, GPT-5.6 Luna, Terra, and Sol, GPT-6 Luna, Sol, and Astra, GPT-6.1 Terra and Sol, gpt-oss-20B and 120B), Grok-4.6 and 4.7, the Gemini series (2.5-Flash-Lite, 2.5-Flash, 2.5-Pro, 3-Flash-Preview, 3-Pro-Preview, 3.1-Flash-Lite-Preview), Qwen3.5 (9B, 122B-A10B) and Qwen3.6 (27B, 35B-A3B, Plus), Gemma-4 (26B-A4B, 31B), DeepSeek-V4 (Flash, Pro), GLM-5 and 5.1, Kimi-K2.5, K2.6, and K3, MiMo-V2.5 and MiMo-V2.5-Pro, and MiniMax-M2.5 and M2.7. A small number of the DeepSeek, GLM, Kimi-K2, MiMo, and MiniMax checkpoints have between 1 and 7 missing users under particular conditions and are annotated accordingly in the appendix. Each model is evaluated under all five memory-ladder conditions on the 42 retained users. Proactive Help Prediction and Emotion-Aware Modeling are reported for the ten Qwen, Gemma, and GPT-5 checkpoints in \cref{tab:proactive,tab:emotion}. The S1 and S2 Cross-Domain Transfer ablations are conducted on the three GPT-5 sizes, yielding 1{,}764 runs in total ($7\,\text{ablations} \times 42\,\text{users} \times 3\,\text{models} \times 2\,\text{conditions}$), and the S1 diagonal is reproduced on Qwen3.6-27B and Gemma-4-31B (\cref{sec:app-val-cdt}).

\section{Results and Analysis}
\label{sec:result}
\subsection{Implicit Preference Reasoning: the user-memory ladder}
\label{sec:results}

\begin{figure}[!t]
\centering
\vspace{-6pt}
\includegraphics[width=\linewidth]{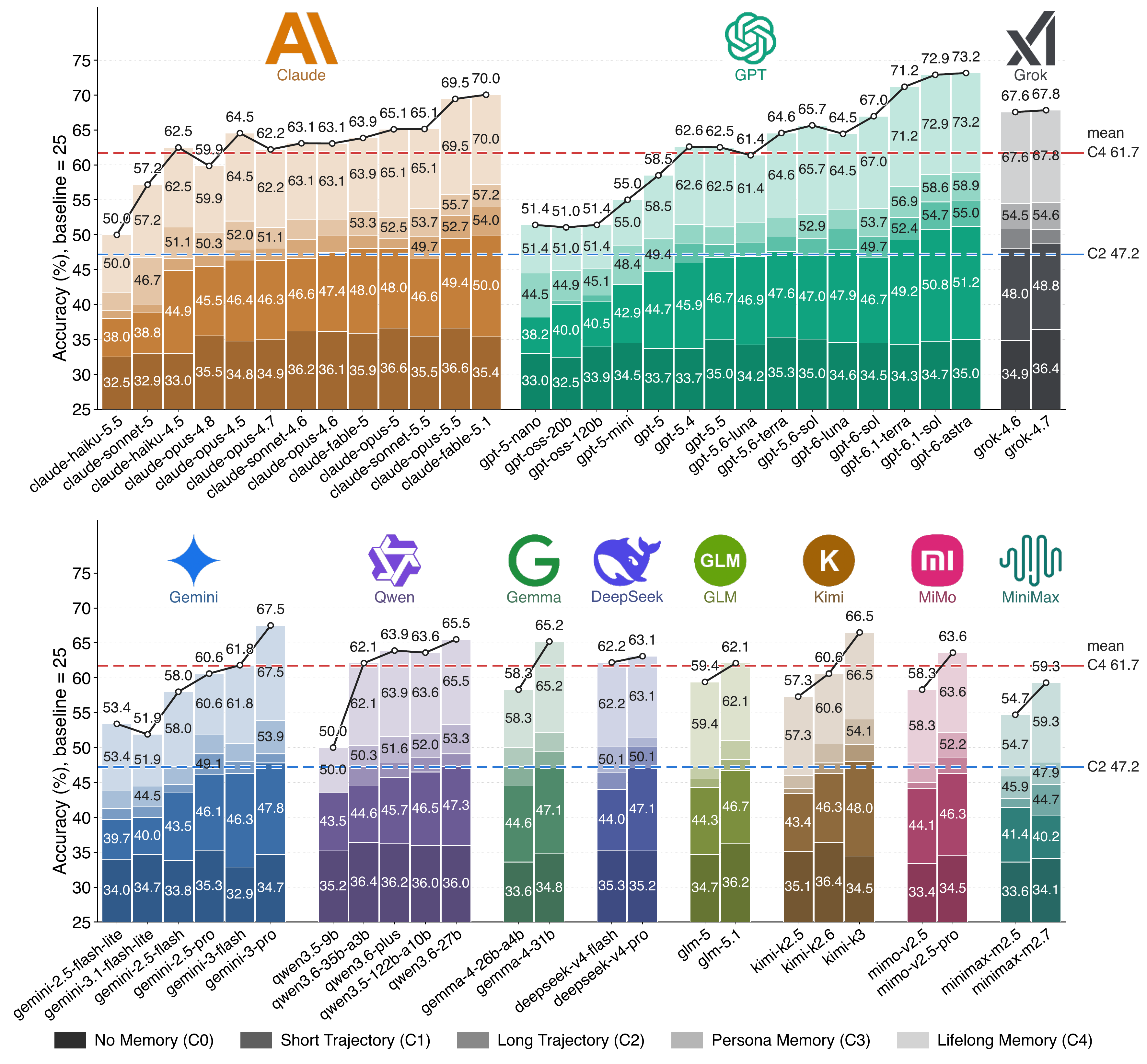}
\caption{{Implicit Preference Reasoning accuracy across memory conditions.} Accuracy across the memory ladder: \textbf{C0} = No Memory, \textbf{C1} = Short Trajectory, \textbf{C2} = Long Trajectory, \textbf{C3} = Persona Memory, \textbf{C4} = Lifelong Memory. Models are grouped by family. Within each family, models are ordered by accuracy, with release date as a small secondary factor, so the black line shows the trend inside the family. The black line traces C4 within each family, and the red and blue dashed lines mark the mean C4 and C2 accuracy over all 54 models.}
\label{fig:main}
\end{figure}

\Cref{fig:main} reports Implicit Preference accuracy across the five-level user-memory ladder for all 54 models; the per-domain breakdowns, the clue-difficulty / clue-type / planted-vs-unplanted stratifications, and the full accuracy matrices (\cref{tab:matrix-a,tab:matrix-b}) are in \cref{app:ipq}. \textsc{Avg}, the equal-weight mean of the five conditions, is a single-figure summary of overall ladder performance.

\definecolor{cBest}{HTML}{059669}        
\definecolor{cGroup}{HTML}{F3F4F6}       
\definecolor{cUp}{HTML}{FEF2F2}          
\definecolor{cDown}{HTML}{EFF6FF}        
\definecolor{bandQwenLite}{HTML}{F4F8F2}    
\definecolor{bandGemmaLite}{HTML}{FAF6F8}   
\definecolor{bandGPTLite}{HTML}{F8F4EE}     
\newcommand{\best}[1]{\textcolor{cBest}{\textbf{#1}}}
\newcommand{\up}[1]{\cellcolor{cUp}#1}
\newcommand{\down}[1]{\cellcolor{cDown}#1}
\newcommand{\swatchUp}{\fcolorbox{cUp}{cUp}{\rule{0pt}{1.5ex}\hspace{1ex}}}
\newcommand{\swatchDown}{\fcolorbox{cDown}{cDown}{\rule{0pt}{1.5ex}\hspace{1ex}}}
\begin{table}[t]
\centering
\caption{\textbf{Proactive Help Prediction.}
Headline F1 and rationale match rate (RMR) across the ten evaluated checkpoints.
\textbf{U} = per-user mean ($n{=}15$), \textbf{P} = pooled anchors ($n{=}254$, $n_{\text{pos}}{=}111$). $\boldsymbol{\Delta}$ is Image~$-$~Text per metric: \swatchUp{}~light red = image helps, \swatchDown{}~light blue = image hurts.
All values in \%; \best{green} marks the best value per column.}
\label{tab:proactive}
\footnotesize
\setlength{\tabcolsep}{3pt}
\renewcommand{\arraystretch}{1.05}
\resizebox{\textwidth}{!}{%
\begin{tabular}{@{} l c c c c c c c c c c c c @{}}
\toprule
\textbf{Model}
  & \multicolumn{4}{c}{\textbf{Text}}
  & \multicolumn{4}{c}{\textbf{Image}}
  & \multicolumn{4}{c}{$\boldsymbol{\Delta}$ \textbf{(Image$-$Text)}} \\
\cmidrule(lr){2-5}
\cmidrule(lr){6-9}
\cmidrule(l){10-13}
& \textbf{U-F1} & \textbf{U-RMR} & \textbf{P-F1} & \textbf{P-RMR}
& \textbf{U-F1} & \textbf{U-RMR} & \textbf{P-F1} & \textbf{P-RMR}
& \textbf{U-F1} & \textbf{U-RMR} & \textbf{P-F1} & \textbf{P-RMR} \\
\midrule

\rowcolor{bandQwenLite}
\multicolumn{13}{@{}l}{\textit{Qwen family}} \\
\quad Qwen3.6-27B
  & 42.8 & 11.6 & 60.7 & 14.4
  & 24.0 & 6.0 & 60.0 & 18.9
  & \down{$-18.8$} & \down{$-5.6$} & \down{$-0.7$} & \up{$+4.5$} \\
\quad Qwen3.5-122B-A10B
  & 28.0 & 9.5 & 64.2 & 21.6
  & 36.5 & 11.4 & 45.3 & 11.7
  & \up{$+8.5$} & \up{$+1.9$} & \down{$-18.9$} & \down{$-9.9$} \\
\quad Qwen3.6-Plus
  & 26.8 & 6.9 & 66.4 & 13.5
  & 41.9 & 7.3 & 35.3 & 6.3
  & \up{$+15.1$} & \up{$+0.4$} & \down{$-31.1$} & \down{$-7.2$} \\
\quad Qwen3.6-35B-A3B
  & 45.1 & \best{21.9} & \best{70.5} & 27.0
  & 40.4 & 8.2 & 56.0 & 20.7
  & \down{$-4.7$} & \down{$-13.7$} & \down{$-14.5$} & \down{$-6.3$} \\
\quad Qwen3.5-9B
  & 23.2 & 8.8 & 60.7 & 30.6
  & 5.2 & 1.4 & 17.5 & 8.1
  & \down{$-18.0$} & \down{$-7.4$} & \down{$-43.2$} & \down{$-22.5$} \\

\rowcolor{bandGemmaLite}
\multicolumn{13}{@{}l}{\textit{Gemma family}} \\
\quad Gemma-4-31B
  & 40.2 & 11.1 & 66.9 & 20.7
  & 36.2 & 9.9 & 64.9 & 31.5
  & \down{$-4.0$} & \down{$-1.2$} & \down{$-2.0$} & \up{$+10.8$} \\
\quad Gemma-4-26B-A4B
  & 33.6 & 5.6 & 62.0 & 18.0
  & 39.9 & 10.5 & 64.8 & 20.7
  & \up{$+6.3$} & \up{$+4.9$} & \up{$+2.8$} & \up{$+2.7$} \\

\rowcolor{bandGPTLite}
\multicolumn{13}{@{}l}{\textit{GPT-5 family}} \\
\quad GPT-5
  & 39.4 & 19.7 & 64.3 & \best{38.7}
  & 52.0 & 14.5 & 61.7 & 36.0
  & \up{$+12.6$} & \down{$-5.2$} & \down{$-2.6$} & \down{$-2.7$} \\
\quad GPT-5-Mini
  & 38.3 & 11.1 & 60.3 & 22.5
  & \best{63.9} & \best{17.7} & \best{65.8} & \best{39.6}
  & \up{\best{$+25.6$}} & \up{\best{$+6.6$}} & \up{\best{$+5.5$}} & \up{\best{$+17.1$}} \\
\quad GPT-5-Nano
  & \best{52.1} & 14.5 & 53.8 & 25.2
  & 49.7 & 10.6 & 58.0 & 29.7
  & \down{$-2.4$} & \down{$-3.9$} & \up{$+4.2$} & \up{$+4.5$} \\
\bottomrule
\end{tabular}%
}
\end{table}

\textbf{Finding 1: Trajectory density is the dominant single contributor.}
Averaged across the 54 models in \cref{fig:main}, the C0-to-C2 step lifts accuracy by about twelve absolute points (from 34.8\% to 47.2\%), whereas adding the persona profile on top (C2 to C3) contributes only a further 3.5 points. The behavioural trace, not the persona profile, drives most of the personalisation gain.

\textbf{Finding 2: A persistent Lifelong-Memory ceiling near 73\%.}
Even under Lifelong Memory (C4), where the full trajectory, persona, and a distilled survey-insight oracle are all available, the strongest model (GPT-6-Astra) reaches only $73.2\%$, leaving more than a quarter of survey questions incorrect. GPT-6.1-Sol ($72.9\%$) is within sampling noise of it, and the best non-GPT model is Claude-Fable-5.1 ($70.0\%$). Implicit-preference inference over real lived behaviour is therefore far from saturated by current frontier LLMs, including the newest GPT and Claude releases.

\textbf{Finding 3: Tier drives gains within a release, and generation usually helps.}
Within a release, larger tiers score higher. In the GPT family, Lifelong-Memory accuracy rises from about 51\% for the smallest checkpoints (gpt-oss-20B, GPT-5-Nano) to $73.2\%$ for GPT-6-Astra, and within the GPT-6 release it grows with tier (Luna $64.5\%$, Sol $67.0\%$, Astra $73.2\%$). Claude 5.5 follows the same order, with Haiku, Sonnet, and Opus at $50.0\%$, $65.1\%$, and $69.5\%$. Newer generations usually help, as GPT-5.4 beats GPT-5 ($62.6\%$ vs.\ $58.5\%$), but not always. Claude-Opus falls from $64.5\%$ (4.5) to $59.9\%$ (4.8), and Claude-Haiku-5.5 trails Claude-Haiku-4.5 ($62.5\%$). Raw parameter count remains a weak predictor across families. Gemma-4-31B (\textsc{Avg} 49.7\%) and Qwen3.6-27B (50.3\%) surpass GPT-5 (45.9\%), Qwen3.5-122B-A10B does not beat its 27B sibling, and small Qwen3.5-9B collapses at Long Trajectory. \textbf{Model tier within a release, more than size across families, predicts preference-reasoning performance on this benchmark.} The per-domain breakdowns in \cref{app:ipq} show \texttt{work\_study} as easy and \texttt{cross\_domain} as hardest across all families.

\textbf{Finding 4: Synthesis densifies the signal rather than creating it.}
Two checks in \cref{app:validation} test whether these results depend on the synthetic layer. On an anchor-only trajectory, C2 accuracy falls by 2 to 11 points and C4 accuracy by 3 to 6 points for four representative models, so the synthetic events mainly add discriminative evidence around the real anchors (\cref{tab:anchor-only}). Re-synthesising 15 users end to end with Qwen3.6-Plus instead of \texttt{gpt-5-mini} moves C4 accuracy by at most 4.7 points and leaves the ranking of five models unchanged (\cref{tab:resynth}).

\subsection{Proactive Help Prediction (text and image)}
\label{sec:results-proactive}

\begin{table}[t]
\centering
\caption{\textbf{Emotion-Aware Modeling.}
Polarity accuracy (3-class) and mood-F1 (multi-label) across the ten evaluated checkpoints.
\textbf{U} = per-user mean ($n{=}15$), \textbf{P} = pooled events ($n{=}123$). $\boldsymbol{\Delta}$ is Image~$-$~Text per metric: \swatchUp{}~light red = image helps, \swatchDown{}~light blue = image hurts.
All values in \%; \best{green} marks the best value per column.}
\label{tab:emotion}
\footnotesize
\setlength{\tabcolsep}{3pt}
\renewcommand{\arraystretch}{1.05}
\resizebox{\textwidth}{!}{%
\begin{tabular}{@{} l c c c c c c c c c c c c @{}}
\toprule
\textbf{Model}
  & \multicolumn{4}{c}{\textbf{Text}}
  & \multicolumn{4}{c}{\textbf{Image}}
  & \multicolumn{4}{c}{$\boldsymbol{\Delta}$ \textbf{(Image$-$Text)}} \\
\cmidrule(lr){2-5}
\cmidrule(lr){6-9}
\cmidrule(l){10-13}
& \textbf{U-Pol} & \textbf{U-Mood} & \textbf{P-Pol} & \textbf{P-Mood}
& \textbf{U-Pol} & \textbf{U-Mood} & \textbf{P-Pol} & \textbf{P-Mood}
& \textbf{U-Pol} & \textbf{U-Mood} & \textbf{P-Pol} & \textbf{P-Mood} \\
\midrule

\rowcolor{bandQwenLite}
\multicolumn{13}{@{}l}{\textit{Qwen family}} \\
\quad Qwen3.6-27B
  & 74.2 & 34.6 & 65.0 & 37.2
  & \best{77.4} & \best{41.0} & 65.0 & \best{41.1}
  & \up{$+3.2$} & \up{\best{$+6.4$}} & $+0.0$ & \up{\best{$+3.9$}} \\
\quad Qwen3.5-122B-A10B
  & 70.8 & 33.6 & 65.9 & \best{39.1}
  & 75.4 & 39.2 & \best{66.7} & 39.5
  & \up{$+4.6$} & \up{$+5.6$} & \up{$+0.8$} & \up{$+0.4$} \\
\quad Qwen3.6-Plus
  & 73.2 & 32.6 & 68.3 & 36.2
  & 73.2 & 34.6 & 62.6 & 36.4
  & $+0.0$ & \up{$+2.0$} & \down{$-5.7$} & \up{$+0.2$} \\
\quad Qwen3.6-35B-A3B
  & 69.1 & 29.2 & 64.2 & 33.4
  & 71.1 & 33.3 & 54.5 & 32.4
  & \up{$+2.0$} & \up{$+4.1$} & \down{$-9.7$} & \down{$-1.0$} \\
\quad Qwen3.5-9B
  & 63.7 & 34.0 & 56.9 & 36.3
  & 42.9 & 19.1 & 33.3 & 17.2
  & \down{$-20.8$} & \down{$-14.9$} & \down{$-23.6$} & \down{$-19.1$} \\

\rowcolor{bandGemmaLite}
\multicolumn{13}{@{}l}{\textit{Gemma family}} \\
\quad Gemma-4-31B
  & 70.5 & 30.9 & 63.4 & 33.6
  & 76.9 & 35.5 & 64.2 & 35.7
  & \up{$+6.4$} & \up{$+4.6$} & \up{$+0.8$} & \up{$+2.1$} \\
\quad Gemma-4-26B-A4B
  & 66.3 & 30.7 & 56.9 & 32.6
  & 75.1 & 30.0 & 62.6 & 28.2
  & \up{\best{$+8.8$}} & \down{$-0.7$} & \up{\best{$+5.7$}} & \down{$-4.4$} \\

\rowcolor{bandGPTLite}
\multicolumn{13}{@{}l}{\textit{GPT-5 family}} \\
\quad GPT-5
  & \best{76.9} & \best{36.2} & \best{69.9} & \best{39.1}
  & 76.5 & 33.3 & 63.4 & 37.3
  & \down{$-0.4$} & \down{$-2.9$} & \down{$-6.5$} & \down{$-1.8$} \\
\quad GPT-5-Mini
  & 67.0 & 25.8 & 60.2 & 31.2
  & 70.7 & 29.4 & 64.2 & 34.2
  & \up{$+3.7$} & \up{$+3.6$} & \up{$+4.0$} & \up{$+3.0$} \\
\quad GPT-5-Nano
  & 76.2 & 30.9 & 63.4 & 36.6
  & 76.5 & 33.4 & 63.4 & 34.6
  & \up{$+0.3$} & \up{$+2.5$} & $+0.0$ & \down{$-2.0$} \\
\bottomrule
\end{tabular}%
}
\end{table}

\Cref{tab:proactive} reports the headline F1 and rationale match rate (RMR) on the proactive task for the ten Qwen, Gemma, and GPT-5 checkpoints, dual-reported as per-user mean (n=15 covered users) and pool micro (n=254 anchors with $n_{\text{pos}}=111$).

\definecolor{cBest}{HTML}{059669}        
\definecolor{cGroup}{HTML}{F3F4F6}       

\textbf{Finding 1: Per-user mean and pool tell different stories.}
At the pool level, Qwen3.6-35B-A3B sits at the top of \cref{tab:proactive} with a text-only F1 of 70.5\%, but its per-user mean is only 45.1\%, more than twenty points lower. The GPT-5 family illustrates the opposite distortion. GPT-5-Nano achieves the highest per-user F1 of 52.1\% in the GPT slice, yet its pool F1 is only 53.8\% because it over-triggers on a small number of anchor-heavy users and is rewarded disproportionately by the per-user macro-average. Single-pool reporting hides this anchor-rich versus anchor-sparse asymmetry, which is why we adopt a dual-reporting protocol on every heterogeneous-coverage task surface in \textsc{Jarvis-Bench}.

\textbf{Finding 2: Multimodality hurts Proactive Help on the Qwen family but rescues parts of the GPT family.}
Comparing the text and image halves of \cref{tab:proactive} for the same model reveals a sharply non-uniform effect of vision. On the pool metric, every Qwen entry except Qwen3.6-27B drops noticeably once the anchor photograph is added: Qwen3.5-9B loses more than forty absolute points of pool F1, Qwen3.6-Plus loses about thirty, and Qwen3.5-122B-A10B loses about nineteen. Qwen3.6-27B itself is hit on the per-user metric, falling from 42.8\% on text to 24.0\% on image. The GPT-5 family behaves in the opposite direction. Image hurts pool F1 only modestly for GPT-5 itself, while raising per-user F1 substantially for both GPT-5 and GPT-5-Mini, with the latter improving by more than twenty-five absolute points when the photograph is added. The mechanism is structural rather than perceptual. Emotion is scene-judgable because the relevant cue lives inside the current event, whereas proactive help is a timing-conditioned trigger that depends on upstream condition chains. Text trajectories compress those chains into a single readable history, but a sequence of timeline photographs dilutes them. For the proactive task, the decisive variable is therefore information compression, not visual richness. Three-run reproductions with an exact paired permutation test at the user level confirm that the image-versus-text gaps of Qwen3.6-27B, Gemma-4-31B, and GPT-5 are significant ($p \leq 0.003$; \cref{tab:proactive-sig}).

\begin{figure}[!t]
\centering
\includegraphics[width=0.96\linewidth]{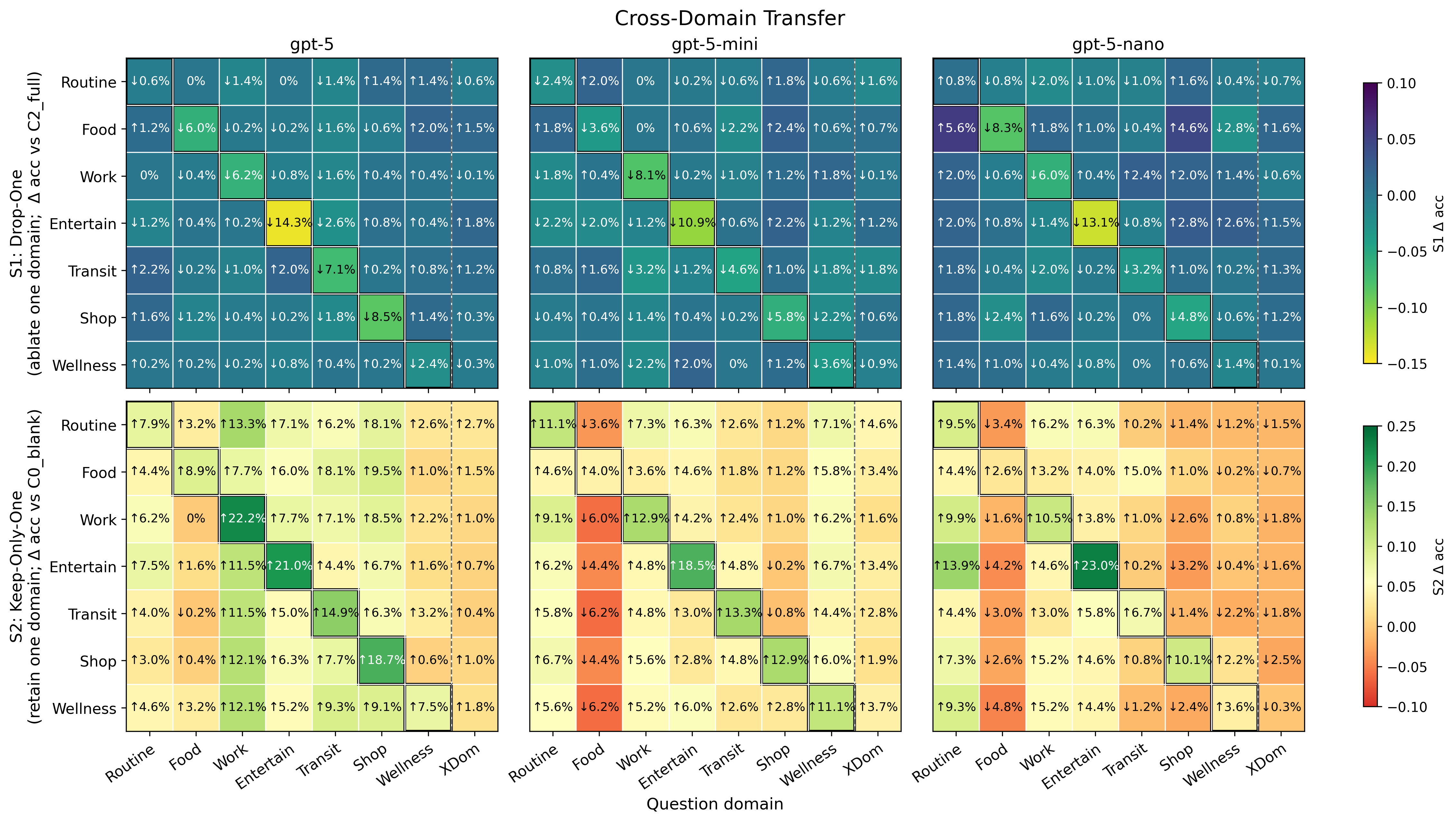}
\caption{\textbf{Cross-Domain Transfer diagonal.} S1 (drop-one, top): negative cells = the dropped lifestyle domain is \emph{necessary} to answer same-domain questions; the larger the negative cell, the more necessary. S2 (keep-only, bottom): positive cells = keeping only that lifestyle domain is \emph{sufficient} for same-domain questions. Panels: GPT-5, GPT-5-Mini, GPT-5-Nano. \texttt{leisure} is simultaneously the most necessary and the most self-sufficient domain.}
\label{fig:domain-transfer}
\end{figure}

\subsection{Emotion-Aware Modeling (text and image)}
\label{sec:results-emotion}

\Cref{tab:emotion} reports the headline polarity accuracy and mood-F1 on the emotion task for the same ten checkpoints. Pool is over n=123 affect-bearing events; per-user is n=15.

\textbf{Finding 1: GPT leads emotion-text; Qwen leads emotion-image.}
GPT-5 takes pool polarity 69.9\% and mood-F1 39.1\% on text (the GPT-5 family also leads the image half of the proactive task, where GPT-5-Mini is best in every column of \cref{tab:proactive}). On image, Qwen3.6-27B leads three of four metrics (per-user F1 41.0\%, pool F1 41.1\%). The image variant matches or surpasses text on per-user polarity in 8 of 10 rows, consistent with the emotion task being scene-judgable.

\textbf{Finding 2: Polarity is partly saturated; mood-F1 is not.}
3-class polarity is partly saturated (top rows in the 63\% to 77\% range); the harder evaluation is multi-label mood, where pool F1 ceilings at 41.1\% (Qwen3.6-27B image). \textbf{Future evaluations should treat mood-F1, not polarity, as the primary metric for the emotion task.}

\subsection{Cross-Domain Transfer}
\label{sec:results-transfer}
The Cross-Domain Transfer task answers a question that the other three tasks cannot: do preferences in different lifestyle domains transfer across one another, or are they silos? We measure transfer with the symmetric S1 / S2 ablations, run on the three GPT-5 sizes (1{,}764 runs total), with the S1 diagonal reproduced on Qwen3.6-27B and Gemma-4-31B. The S1 / S2 conditions act on the trajectory side by selecting events through their event subdomain (per the canonical mapping in \cref{tab:domain-mapping}). \Cref{fig:domain-transfer} visualises both diagonals as percent lifts; per-model breakdowns are in \cref{app:cdt}.

\textbf{Finding 1: \texttt{leisure} is the dominant transfer source.}
On the diagonal, the \texttt{leisure} lifestyle domain carries an S1 drop-one effect of $\Delta = -12.8$ averaged across the three GPT-5 sizes, the most negative entry in the matrix, indicating that it is the most \emph{necessary} domain for answering same-domain questions. The corresponding S2 keep-only effect is $\Delta = +20.8$, the most positive entry, indicating that it is also the most \emph{self-sufficient} domain. Removing every \texttt{entertainment} event from the trajectory costs the three models between eleven and fourteen absolute points on leisure questions. The S1 diagonal of Qwen3.6-27B and Gemma-4-31B again makes \texttt{leisure} the most negative entry ($\Delta = -15.5$ and $-10.5$), with \texttt{work\_study} again second (\cref{tab:cdt-families}). 

\textbf{Finding 2: Most lifestyle domains are individually informative.}
The S2 keep-only diagonal $\Delta$ is positive on all seven ablation domains, and four of them, namely \texttt{commuting\_travel}, \texttt{leisure}, \texttt{shopping}, and \texttt{work\_study}, lift same-domain accuracy by more than ten absolute points on their own. After adding back the events of a single lifestyle domain, the three GPT-5 sizes reach roughly 37\% to 50\% same-domain accuracy on the diagonal, starting from a No-Memory baseline that ranges from about 27\% to 42\% across domains. 

\textbf{Finding 3: Smaller models extract less from a single lifestyle domain.}
Across the three GPT-5 sizes, the S2 keep-only $\Delta$ on \texttt{food\_dining} drops monotonically with model size, from $+8.9$ on the full GPT-5, to $+4.0$ on the Mini variant, and down to $+2.6$ on the Nano variant. Smaller models extract less information from a single lifestyle domain in isolation and depend more heavily on the full multi-domain context to recover comparable accuracy. On this benchmark, smaller models are therefore \emph{less} efficient per unit of preference-bearing context, which inverts the common intuition that smaller models economise on tokens.


\FloatBarrier
\section{Limitations}
\label{sec:discussion}

\textsc{Jarvis-Bench} has several limitations as follows.
(i) \textbf{Scale.} 70 enrolled / 42 evaluated is modest. Of the 28 excluded participants, 14 logged too little, 6 left the survey incomplete, and 8 failed the fidelity confirmation, an attrition rate of 40\%. Each retained user is densely covered (18,471 evaluated events and 4,200 survey answers in total), but the underlying preference distribution is still that of 42 humans; a larger recruitment round through our data-collection app and participant platform is in preparation.
(ii) \textbf{Population scope.} Participants are concentrated in the United States (75\%), with smaller groups from Southeast Asia, Europe, and China (\cref{fig:participant-diversity}). Our conclusions, including the gap to robust personal intelligence, hold within this population, and preferences may not generalise globally for the food, leisure, and social lifestyle domains.
(iii) \textbf{Generator-tier dependence.} The synthesis stages use \texttt{gpt-4o} and \texttt{gpt-5-mini}, both from a single vendor. Re-synthesising 15 users with Qwen3.6-Plus preserves the model ranking with C4 shifts of at most 4.7 points (\cref{sec:app-val-synth}), but a full re-synthesis with other generators is left to future work.
(iv) \textbf{No third-party human baseline.} Every label is the participant's own answer, so the participant defines a 100\% ceiling by construction. We do not report an outside annotator's accuracy, because a stranger cannot know a person's implicit preferences better than the person does.

\section{Conclusion}
\label{sec:conclusion}
We introduced \textsc{Jarvis-Bench}, a benchmark for personal intelligence agents grounded in 70 enrolled and 42 qualified real participants who provided 28 to 44 days of multimodal daily traces, persona profiles, longitudinal feedback, and a 100-question implicit-preference survey. The synthetic-trajectory layer enriches the real anchor logs without replacing them, and the four task surfaces (Implicit Preference Reasoning, Proactive Help Prediction, Emotion-Aware Modeling, Cross-Domain Transfer) consume the same artefact through one iterator. Across 54 LLMs from 11 families, even the strongest model (GPT-6-Astra) answers only 73.2\% of survey questions correctly under oracle context, no single family wins all four task surfaces, and the Cross-Domain Transfer ablation reveals \texttt{leisure} as the dominant transfer source in all three model families tested. These conclusions hold within the population covered by the dataset. 

\begin{ack}
We thank all study participants for their time, sustained engagement, and valuable contributions to this research. We also thank the BDSC group at the University of Illinois Chicago and everyone who provided helpful feedback and suggestions that improved this work.

\end{ack}

\bibliographystyle{apalike}
\bibliography{references}


\newpage

\appendix

\FloatBarrier
\section{Extended Related Work}
\label{app:related-full}

\begin{table}[htpb]
    \centering
    \small
    \renewcommand{\arraystretch}{0.96}
    \setlength{\tabcolsep}{2.8pt}
    \caption{Comparison with representative personalization, lifelog, and proactive/emotional benchmarks.  \textsc{Jarvis-Bench} is the only benchmark grounded in multi-user real recorded life with survey-anchored implicit preference labels and participant-self-recorded help-need flags and emotion annotations. \textbf{Real Traj.} is judged at the data-collection level. A check means that the benchmark's behavioural data and labels come from long-term records of real users rather than short scripted sessions or simulated users. In \textsc{Jarvis-Bench} every anchor event and every label is real, and synthetic events only densify the trajectory between anchors.}
    \label{tab:benchmark_comparison_updated}
    \begin{tabular}{@{}p{0.085\textwidth} p{0.23\textwidth} c c c c c c c c@{}}
        \toprule
        \textbf{Data} & \textbf{Benchmark} & \textbf{Real} & \textbf{Real} & \textbf{Cross-} & \textbf{Implicit} & \textbf{Multi-} & \textbf{Multi-} & \textbf{Proact.} & \textbf{Emotion} \\
        \textbf{Source} & & \textbf{Profile} & \textbf{Traj.} & \textbf{Domain} & \textbf{Pref.} & \textbf{modal} & \textbf{lingual} & & \\
        \midrule
        \multirow{12}{*}{Synthetic}
        & TimelineQA \citep{tan2023timelineqa}          & \redcross   & \redcross   & \greencheck & \redcross   & \redcross   & \redcross   & \redcross   & \redcross   \\
        & LoCoMo \citep{maharana2024evaluating}         & \redcross   & \redcross   & \redcross   & \redcross   & \greencheck & \redcross   & \redcross   & \redcross   \\
        & PrefEval \citep{zhao2025llms}                 & \redcross   & \redcross   & \halfcheck  & \greencheck & \redcross   & \redcross   & \redcross   & \redcross   \\
        & PersonaBench \citep{tan2025personabench} & \redcross & \redcross & \greencheck & \halfcheck & \redcross & \redcross & \redcross & \redcross \\
        & PersonaMem \citep{jiang2025know}              & \redcross   & \redcross   & \greencheck & \halfcheck  & \redcross   & \redcross   & \redcross   & \redcross   \\
        & PersonaMem-v2 \citep{jiang2025personamem} & \redcross & \redcross & \greencheck & \halfcheck & \greencheck & \greencheck & \redcross & \redcross \\
        & LifeBench \citep{cheng2026lifebench}          & \redcross   & \redcross   & \greencheck & \halfcheck  & \greencheck & \redcross   & \redcross   & \redcross   \\
        & LifeSim-Eval \citep{duan2026lifesim}          & \redcross   & \redcross   & \greencheck & \greencheck & \redcross   & \redcross   & \redcross   & \redcross   \\
        & PSPA-Bench \citep{nie2026pspa} & \redcross & \redcross & \halfcheck & \halfcheck & \greencheck & \redcross & \redcross & \redcross \\
        & ASTRA-Bench \citep{xiu2026astra} & \redcross & \redcross & \halfcheck & \halfcheck & \greencheck & \greencheck & \redcross & \redcross \\
        & ESL-Bench \citep{li2026eslbench}              & \redcross   & \redcross   & \redcross   & \redcross   & \greencheck & \redcross   & \redcross   & \redcross   \\
        & PERMA \citep{liu2026perma}                    & \redcross   & \redcross   & \greencheck & \greencheck & \redcross   & \redcross   & \redcross   & \redcross   \\
        & ES-MemEval \citep{chen2026esmemeval}          & \redcross   & \redcross   & \redcross   & \halfcheck  & \redcross   & \redcross   & \redcross   & \greencheck \\
        \midrule
        \multirow{14}{*}{Real}
        & PersonaChat \citep{zhang2018personalizing}    & \redcross  & \redcross   & \redcross   & \redcross   & \redcross  & \redcross   & \redcross   & \redcross   \\
        & REALTALK \citep{lee2025realtalk} & \redcross & \greencheck & \redcross & \halfcheck & \redcross & \redcross & \redcross & \halfcheck \\
        & ESConv \citep{liu2021esconv}                  & \redcross  & \redcross   & \redcross   & \redcross   & \redcross  & \redcross   & \redcross   & \greencheck \\
        & LaMP \citep{salemi2024lamp}                   & \redcross  & \greencheck & \redcross   & \halfcheck  & \redcross  & \redcross   & \redcross   & \redcross   \\
        & MemoryCD \citep{zhang2026memorycd}            & \redcross  & \greencheck & \halfcheck   & \greencheck & \redcross  & \redcross   & \redcross   & \redcross   \\
        & AlpsBench \citep{xiao2026alpsbench}           & \redcross  & \greencheck & \greencheck & \greencheck & \redcross  & \redcross   & \redcross   & \redcross   \\
        & OpenLifelogQA \citep{tran2025openlifelogqa}   & \redcross  & \greencheck & \greencheck & \redcross   & \greencheck& \redcross   & \redcross   & \redcross   \\
        & ETRI Lifelog \citep{oh2025understanding}      & \halfcheck & \greencheck & \halfcheck  & \redcross   & \greencheck& \redcross   & \redcross   & \halfcheck  \\
        & ATM-Bench \citep{mei2026according}            & \redcross  & \greencheck & \greencheck & \redcross   & \greencheck& \redcross   & \redcross   & \redcross   \\
        & KnowMe-Bench \citep{wu2026knowme}             & \halfcheck & \redcross   & \greencheck & \greencheck & \redcross & \redcross   & \redcross   & \halfcheck  \\
        & LifeAgentBench \citep{tian2026lifeagentbench} & \redcross  & \greencheck & \halfcheck   & \redcross   & \greencheck& \redcross   & \redcross   & \halfcheck  \\
        & ProactiveBench \citep{lu2024proactiveagent}   & \redcross  & \halfcheck  & \halfcheck  & \redcross   & \greencheck& \redcross   & \greencheck & \redcross   \\
        & ProAgentBench \citep{tang2026proagentbench}   & \redcross  & \greencheck & \halfcheck  & \redcross   & \greencheck& \redcross   & \greencheck & \redcross   \\
        \cline{2-10}
        & \mbox{\textbf{\textsc{Jarvis-Bench} (Ours)}} & \greencheck & \greencheck & \greencheck & \greencheck & \greencheck & \greencheck & \greencheck & \greencheck \\
        \bottomrule
    \end{tabular}
    \begin{tablenotes}
    \footnotesize
    \item \greencheck: fully covered, \halfcheck: partially covered, \redcross: not covered. 
    \item \textbf{Real Profile}: user profile content is sourced \textit{from} the participant (e.g., self-reported via survey, demographics released as part of the benchmark), as opposed to crowdworker-authored, LLM-generated, or assigned persona cards.
    \item \textbf{Real Trajectory}: judged at the data-collection level, that is, whether the benchmark's behavioural records and labels are derived from long-term records of participants' actual lived behavior, as opposed to crowdworker role-play, simulator-generated users, or short scripted sessions. Benchmarks whose real trajectories last only minutes per user receive a half check.
    \item \textbf{Cross-Domain}: covers at least three broad everyday-life domains beyond a single application vertical, such as food, media, exercise, work, transit, social, shopping, app-use, or travel.
    \item \textbf{Implicit Preference}: task requires inferring user preferences from behavioral or interaction evidence, rather than retrieving explicit persona statements or directly stated preferences.
    \item \textbf{Multimodal}: benchmark includes multiple data modalities beyond plain text (e.g., images, sensors, wearables).
    \item \textbf{Multilingual}: benchmark natively contains interaction data, instructions, queries, or annotations in two or more languages, rather than only being machine-translated by users after release.
    \item \textbf{Proactive}: includes participant-self-recorded help-need flags within the trajectory.
    \item \textbf{Emotion}: includes participant-self-reported in-day emotion annotations.
    \end{tablenotes}
\end{table}

We summarise prior work along four axes that map directly onto the four properties \textsc{Jarvis-Bench} jointly satisfies: real-user grounding, longitudinal multi-domain coverage, behavioural-implicit supervision, and joint evaluation across preference, proactive, and emotional capabilities.

\paragraph{Persona-grounded LLM evaluation.}
Long-running benchmarks evaluate whether dialogue models and LLM agents can condition on explicit personas, textual profiles, private documents, or remembered interaction histories, and then generate responses consistent with those signals~\citep{zhang2018personalizing, zheng2019personalized, maharana2024evaluating, tan2025personabench, jiang2025know, tao2025personafeedback, lee2025realtalk, salemi2024lamp}. These works are valuable for measuring persona consistency, profile retrieval, and personalized response generation, but the relevant user information is usually already expressed in language as persona sentences, profile attributes, dialogue facts, private documents, or explicit feedback~\citep{zhang2025toward}. This makes the evaluation closer to memory access or profile application than to discovering preferences that were never directly stated. PrefEval~\citep{zhao2025llms} and KnowMe-Bench~\citep{wu2026knowme} move toward implicit-preference and person-understanding tasks, and MemoryCD~\citep{zhang2026memorycd} grounds personalization in authentic Amazon review histories. MemAudit~\citep{ma2026memaudit} audits whether an agent's long-term memory recovers a hidden user state. PersonaAgent~\citep{zhang2026personaagent} couples a personalized memory module with user-tailored tool actions, memory gating can be learned per user~\citep{ji2026learning}, and surveys and harness evaluations of agent memory study how agents represent, store, and reuse information over long horizons~\citep{huang2026rethinking,huang2026harness,guo2026seeing}. However, these settings still rely on textual conversations, autobiographical narratives, or platform-bounded consumption traces. \textsc{Jarvis-Bench} instead grounds users in recorded daily trajectories and requires models to infer preferences from behavioural consequences, routines, avoided options, cross-domain habits, and help-seeking signals, while explicitly forbidding survey-answer restatement in the persona.

\paragraph{Lifelog and self-tracking corpora.}
Lifelog and self-tracking datasets preserve real lived experience through wearable cameras, smartphones, sensors, images, emails, videos, and life journals. LLQA~\citep{tran2022llqa}, OpenLifelogQA~\citep{tran2025openlifelogqa}, the LSC collection~\citep{gurrin2019testcollection,gurrin2022lsc}, ETRI Lifelog~\citep{jung2022etri}, UbiqLog~\citep{rawassizadeh2013ubiqlog}, AutoLife~\citep{xu2025autolife}, and ATM-Bench~\citep{mei2026according} support retrieval, recognition, journaling, activity understanding, and memory QA. Yet they are often built around one lifelogger or a small contributor pool, and their tasks typically ask models to recall or answer questions about recorded events rather than infer latent preferences across many users. In addition, raw lifelog streams are sparse and uneven: some domains are well observed, while subjective preferences, avoided choices, and moments of needing help are often missing. Larger authentic traces, such as LifeAgentBench~\citep{tian2026lifeagentbench}, OmniBehavior~\citep{chen2026omnibehavior}, OPeRA~\citep{wang2025opera}, and Customer-R1~\citep{wang2025customer}, increase user scale but remain bounded to health, online platforms, or shopping. Evaluations of personal health agents over users' health memory and sensor metrics~\citep{zhang2026rubricstree} likewise stay within the health vertical. As a result, they are less suited to evaluating a general personal assistant that must reason across everyday domains rather than optimize within one application vertical. \textsc{Jarvis-Bench} occupies the middle ground: it starts from real multi-user self-tracking logs and overlays a dense synthetic skeleton constrained by each participant's persona and survey answers, enabling multi-hop preference inference across eight lifestyle domains rather than isolated retrieval, health QA, or platform-specific behaviour simulation.

\paragraph{Synthetic behaviour and trajectory generation.}
LLM-driven user simulators and synthetic agents generate plausible behaviour for evaluation and training. Generative Agents~\citep{park2023generative} simulate social behaviour in a sandbox, BehaviorChain~\citep{li2025behaviorchain} evaluates persona-conditioned behaviour chains, and Agent4Rec~\citep{zhang2024agent4rec}, RecAgent~\citep{wang2024recagent}, and USimAgent~\citep{zhang2024usimagent} simulate recommendation or search users. Synthetic personal-life benchmarks, including TimelineQA~\citep{tan2023timelineqa}, MemSim~\citep{zhang2024memsim}, LifeBench~\citep{cheng2026lifebench}, ESL-Bench~\citep{li2026eslbench}, LifeSim~\citep{duan2026lifesim}, and PSPA-Bench~\citep{nie2026pspa}, improve density, controllability, and ground-truth availability through generated users, memories, health trajectories, life events, or GUI instructions. These benchmarks provide clean supervision but their users and labels are produced by simulation or profile-construction pipelines. This matters because synthetic users can be internally coherent while still failing to reflect the irregularity, sparsity, contradictions, and idiosyncrasies of real people. SimBench~\citep{hu2025simbench} and REALTALK~\citep{lee2025realtalk} further caution that plausible synthetic behaviour is not necessarily faithful human behaviour. \textsc{Jarvis-Bench} therefore treats synthesis as an augmentation layer: the seed trajectories come from real participants, and the preference labels come from their own survey answers. The goal is not to simulate a generic believable person, but to densify evidence around a specific real participant without changing that participant's ground-truth preferences.

\paragraph{Proactive and emotional AI agents.}
Proactive-help benchmarks study when agents should intervene and what assistance they should provide. ProactiveBench~\citep{lu2024proactiveagent}, ContextAgentBench~\citep{yang2025contextagent}, FingerTip 20K~\citep{yang2025fingertip20k}, ProAgentBench~\citep{tang2026proagentbench}, ProactiveEval~\citep{liu2025proactiveeval}, and PROBE~\citep{pasternak2025beyondreactivity} evaluate proactive intervention, intent prediction, mobile assistance, workflow support, or problem solving, but are usually scenario-driven, tool-centric, interface-specific, or work-domain focused. They also often define proactivity from the system side, through generated moments or task success, rather than from the participant's own logged sense that help was needed. Work on LLM-based human-agent systems argues for keeping humans in the loop and evaluates agents under configurable human participation~\citep{zou2025survey,zou2025call,wu2026hasbench}, but it studies human involvement within a task rather than help needs that users log in daily life. Emotional-intelligence and support benchmarks, including EmpatheticDialogues~\citep{rashkin2019empatheticdialogues}, ESConv~\citep{liu2021esconv}, ExTES~\citep{zheng2023extes}, EmotionBench~\citep{huang2024emotionbench}, EmoBench~\citep{sabour2024emobench}, EQ-Bench~\citep{paech2023eqbench}, HumDial-EIBench~\citep{wang2026humdial}, HEART~\citep{iyer2026heart}, MentalChat16K~\citep{xu2025mentalchat16k}, ES-MemEval~\citep{chen2026esmemeval}, ETrajEval~\citep{tan2025etrajeval}, A-MBER~\citep{wen2026amber}, and MemEmo~\citep{liu2026mememo}, usually center dialogue, counseling, synthetic scenarios, or affective memory QA. These benchmarks evaluate important affective skills, but they rarely place emotion inside the same behavioural trajectory used for preference inference and proactive assistance. To our knowledge, \textsc{Jarvis-Bench} is the first benchmark to combine participant-self-recorded proactive-help labels with participant-self-reported in-day mood annotations inside a multi-day, multi-domain real-user behavioural trajectory, evaluated alongside Implicit Preference Reasoning and Cross-Domain Transfer.

\FloatBarrier
\section{Dataset Construction (Full Detail)}
\label{app:dataset-construction}
This appendix provides the full detail of how each \textsc{Jarvis-Bench} user artefact is built end-to-end, from participant recruitment through the human-in-the-loop fidelity check that closes the loop.

\subsection{\textsc{Jarvis-Bench} at a glance}
\label{sec:app-summary}

\Cref{tab:dataset-summary} summarises the released benchmark in one place. We recruited 70 IRB-consented participants, who logged around-the-clock through a custom mobile interface. After the three-stage quality bar described in \cref{sec:app-loop}, 42 of the 70 participants qualified for the released evaluation set; the other 28 were excluded due to insufficient logging volume, incomplete survey response, or failure to confirm fidelity at the final stage. All benchmark numbers in this paper are computed on the 42 retained users, who logged 28 to 44 days each and contribute about 440 events per user, roughly half anchor (real) events and half synthetic events that wrap the anchors in coherent context.

\begin{table}[h]
\centering
\caption{\textbf{\textsc{Jarvis-Bench} at a glance.} 70 IRB-consented participants enrolled; the 42 who passed the three-stage quality bar logged 28 to 44 days each and form the released evaluation set. Anchor events preserve the participant's original log content verbatim; synth events are clue-planted expansions whose construction is detailed in \cref{sec:app-pipeline}.}
\label{tab:dataset-summary}
\small
\setlength{\arrayrulewidth}{0.4pt}
\arrayrulecolor{tabHeader}
\begin{tabularx}{\linewidth}{@{}lY@{}}
\toprule
\rowcolor{tabHeader}
\hdrcell{Aspect} & \hdrcell{Value} \\
\midrule
\rowcolor{bandExt}\multicolumn{2}{@{}l@{}}{\,\textbf{\textcolor{tabHeader}{Population and scale}}} \\
Participants (enrolled \& logged / qualified for eval) & \textbf{70 / 42} \\
\rowcolor{rowAlt} Logging window per user & 28 to 44 days, around-the-clock multimodal \\
Events per user (released set) & $\sim$440 (anchor + synth) \\
\rowcolor{rowAlt} Synthetic ratio & $\sim$0.5 (anchors preserved verbatim) \\
\midrule
\rowcolor{bandQual}\multicolumn{2}{@{}l@{}}{\,\textbf{\textcolor{tabHeader}{Domain and time taxonomy}}} \\
Log event subdomains (12) & wake, bedtime, nap, food, work, entertainment, weather, travel, shopping, physical, exercise, event \\
\rowcolor{rowAlt} \textbf{Survey lifestyle domains (8 = 7 lifestyle + 1 cross-domain)} & \texttt{daily\_routine}, \texttt{food\_dining}, \texttt{work\_study}, \texttt{leisure}, \texttt{commuting\_travel}, \texttt{shopping}, \texttt{social\_wellbeing}, \texttt{cross\_domain} \\
Time buckets & 9 (deep\_night to late\_night) \\
\midrule
\rowcolor{bandInt}\multicolumn{2}{@{}l@{}}{\,\textbf{\textcolor{tabHeader}{Multimodal and affect coverage}}} \\
Anchors with photograph & $\sim$40\% \\
\rowcolor{rowAlt} Anchors with mood annotation (\texttt{moodKeywords}/\texttt{moodNote}) & $\sim$38\% \\
Anchors flagged \texttt{agentHelp = true} & $\sim$5\% (used for proactive ground truth) \\
\midrule
\rowcolor{bandExt}\multicolumn{2}{@{}l@{}}{\,\textbf{\textcolor{tabHeader}{Survey instrument and clue plan}}} \\
Survey instrument & 100 multiple-choice questions $\times$ 4 options, 8 buckets \\
\rowcolor{rowAlt} Clue difficulty (planted) & hard, very hard \\
Clue kinds (Stage 3 prompt) & timing-detail, consequence, decision-frame, post-event-mood, social-context, app-choice, tiny-detail \\
\bottomrule
\end{tabularx}
\arrayrulecolor{black}
\end{table}

\subsection{Real-user collection}
\label{sec:app-source}

We recruited \textbf{70 IRB-consented participants} under a self-tracking protocol approved by the Institutional Review Board of the University of Illinois Chicago. Logging ran for 28 to 44 days per retained participant through a custom mobile interface that prompted around-the-clock entry across the 12 event subdomains of \cref{tab:domain-mapping}. Each participant contributed three forms of user-provided supervision:

\begin{itemize}[leftmargin=*, nosep]
\item \textbf{Daily logs} with optional photographs, timing fields, and free-text descriptions, preserved verbatim as anchor events. Embedded within each log entry are two further annotation channels that serve as ground truth for downstream tasks: per-event proactive-help flags accompanied by a free-text rationale (the supervision signal for Proactive Help Prediction), and free-text mood notes together with mood keywords on affect-bearing events (the supervision signal for Emotion-Aware Modeling).
\item \textbf{Persona profile}: structured demographics (age band, gender, region, role) plus a free-text self-description.
\item \textbf{100-question implicit-preference instrument} over 8 lifestyle domains (7 lifestyle plus 1 cross-domain), four options per question. See \cref{sec:app-domain-taxonomy} for the canonical mapping between event subdomains (trajectory side) and lifestyle domains (survey side).
\end{itemize}

After the three-stage quality bar described in \cref{sec:app-loop}, whose final stage is a participant-driven fidelity confirmation on a randomly sampled subset of synthetic events, \textbf{42 of the 70 participants} qualified for the released evaluation set. The remaining 28 were excluded due to insufficient logging volume (14 participants, with windows below 28 days or fewer than the minimum number of structured events), incomplete survey response (6), or failure to confirm fidelity at the final stage (8), an overall attrition rate of 40\%.

\paragraph{Privacy preservation.}
Calendar dates are stripped at ingestion and replaced by relative \texttt{Day\{N\}\allowbreak T\{HH:\allowbreak MM:\allowbreak SS\}} timestamps. Surnames, precise locations, and calendar-day identifiers do not appear in any released artefact. Photographs enter the trajectory text only through action-focused captions written by a vision LLM. A privacy blacklist enforced by the structural validator (\cref{sec:app-loop}) rejects calendar dates, weekday names, and 12-hour-clock times in any synthetic text.

\subsection{Two-level domain taxonomy}
\label{sec:app-domain-taxonomy}

\textsc{Jarvis-Bench} uses two levels of domain labelling, and the distinction is load-bearing for every per-domain analysis in the paper.

\paragraph{Lifestyle domain (high-level, 8 labels) is the survey-side taxonomy.}
The 100-question implicit-preference instrument and every per-domain stratification reported in the main results are organised into \textbf{8 lifestyle domains}: 7 substantive lifestyle areas plus one cross-cutting bucket. The 7 lifestyle domains are \texttt{daily\_routine}, \texttt{food\_dining}, \texttt{work\_study}, \texttt{leisure}, \texttt{commuting\_travel}, \texttt{shopping}, \texttt{social\_wellbeing}; the 8th, \texttt{cross\_domain}, contains questions that span multiple lifestyle areas and is therefore not an ablation target on the trajectory side. We use the term \emph{lifestyle domain} throughout the paper for this 8-label taxonomy, and \emph{cross-domain} for the questions of the eighth label specifically.

\paragraph{Event subdomain (fine-grained, 12 labels plus long-tail) is the trajectory-side taxonomy.}
On the trajectory side, every event in a user's daily log carries a fine-grained \textbf{event subdomain} drawn from the 12 canonical labels \{\texttt{wake}, \texttt{bedtime}, \texttt{nap}, \texttt{food}, \texttt{work}, \texttt{entertainment}, \texttt{weather}, \texttt{travel}, \texttt{shopping}, \texttt{physical}, \texttt{exercise}, \texttt{event}\}, with any additional self-reported subdomain falling into a long-tail \texttt{other} bucket. Each subdomain is a concrete event type the participant can log through the mobile interface (e.g., \texttt{food} for a meal, \texttt{entertainment} for a media-consumption episode). The full released trajectory artefact preserves the participant-reported subdomain on every anchor event, and the synthesis pipeline assigns one of the same subdomain labels to every synthetic event.

\paragraph{Mapping between the two levels.}
\Cref{tab:domain-mapping} reproduces the canonical mapping released with the benchmark. Each lifestyle domain absorbs one or more event subdomains; the mapping is many-to-one and is the single source of truth used by the Cross-Domain Transfer ablation harness and by every per-domain stratification table in the paper. Long-tail events that fall outside the 12 canonical labels are bucket-agnostic and are kept in the trajectory regardless of S1 or S2 ablation.

\begin{table}[h]
\centering
\caption{\textbf{Two-level domain taxonomy.} Survey-side \emph{lifestyle domain} (8 = 7 lifestyle + 1 cross-domain) maps from one or more trajectory-side \emph{event subdomain} labels. Cross-domain has no event subdomains since cross-domain questions are defined to span multiple lifestyle areas. Long-tail \texttt{other} subdomains do not map to any lifestyle domain and are kept in the trajectory regardless of S1 / S2 ablation.}
\label{tab:domain-mapping}
\small
\arrayrulecolor{tabHeader}
\renewcommand{\arraystretch}{1.18}
\begin{tabularx}{\linewidth}{@{}l Y l@{}}
\toprule
\rowcolor{tabHeader}
\hdrcell{Lifestyle domain (survey, 8)} & \hdrcell{Event subdomains (trajectory, 12 + other)} & \hdrcell{S1 / S2 ablation target?} \\
\midrule
\rowcolor{bandExt} \texttt{daily\_routine} & \texttt{wake}, \texttt{bedtime}, \texttt{nap} & \yes \\
\rowcolor{rowAlt} \texttt{food\_dining} & \texttt{food} & \yes \\
\rowcolor{bandExt} \texttt{work\_study} & \texttt{work} & \yes \\
\rowcolor{rowAlt} \texttt{leisure} & \texttt{entertainment} & \yes \\
\rowcolor{bandExt} \texttt{commuting\_travel} & \texttt{weather} (carries \texttt{leaveHomeTime}, \texttt{returnHomeTime}), \texttt{travel} & \yes \\
\rowcolor{rowAlt} \texttt{shopping} & \texttt{shopping} & \yes \\
\rowcolor{bandExt} \texttt{social\_wellbeing} & \texttt{physical}, \texttt{exercise}, \texttt{event} & \yes \\
\midrule
\rowcolor{bandHigh} \texttt{cross\_domain} & (none; spans multiple lifestyle domains) & \no \\
\midrule
\rowcolor{rowAlt} (long-tail) & \texttt{other} (any self-reported subdomain not in the 12 above) & \no (always kept) \\
\bottomrule
\end{tabularx}
\arrayrulecolor{black}
\end{table}

\paragraph{When each term is used in the paper.}
We use \emph{lifestyle domain} (or simply \emph{domain} when context is unambiguous) throughout the main body whenever we refer to the 8-label survey taxonomy: in the per-domain breakdowns (\cref{fig:domain-workstudy,fig:domain-crossdomain}), in the Cross-Domain Transfer findings (\cref{sec:results-transfer}), and in the S1 / S2 ablation IDs (\texttt{drop\_<domain>}, \texttt{keep\_<domain>}). We use \emph{event subdomain} on the trajectory side: in the released log schema, in the composition figures (\cref{fig:composition-events,fig:composition-tokens,fig:composition-per-user}), and in the synthesis pipeline's anchor-fidelity contract. The two levels never interchange: an event has an event subdomain, a survey question has a lifestyle domain, and the canonical mapping above is the only bridge between them.

\subsection{Clue-planted synthesis pipeline}
\label{sec:app-pipeline}

The synthesis pipeline transforms each user's sparse, post-hoc-biased raw logs into a dense, narrative-coherent multi-day trajectory while preserving the original anchor entries verbatim. The four stages and their invariants are summarised in \cref{tab:pipeline}.

\begin{table}[h]
\centering
\caption{\textbf{Synthesis pipeline at a glance.} \texttt{survey\_insight} is the \emph{only} conduit through which raw survey content reaches Stages 2 and 3; the 100-question text is never re-injected.}
\label{tab:pipeline}
\small
\arrayrulecolor{tabHeader}
\renewcommand{\arraystretch}{1.2}
\begin{tabularx}{\linewidth}{@{}>{\raggedright\arraybackslash}p{1.5cm} >{\raggedright\arraybackslash}p{2.2cm} Y >{\raggedright\arraybackslash}p{2.4cm} >{\raggedright\arraybackslash}p{2.85cm}@{}}
\toprule
\rowcolor{tabHeader}
\hdrcell{Stage} & \hdrcell{Role} & \hdrcell{Input $\to$ Output} & \hdrcell{Default model} & \hdrcell{Key invariant} \\
\midrule
\rowcolor{bandExt} 0 & Image grounding & user photographs $\to$ third-person action captions and abstracted captions & vision LLM (\texttt{gpt-4o}) & describe what the user is \emph{doing}, not what is in the frame \\
\rowcolor{rowAlt} 1 & Persona plus survey insight & persona, 100 survey answers, local trait digest, sample captions $\to$ \texttt{enhanced\_profile} (18 axes), \texttt{uniqueness\_\allowbreak signature}, \texttt{persona\_narrative}, \texttt{survey\_insight} & reasoning LLM (\texttt{gpt-5-mini}, low effort) & \texttt{survey\_insight} = sole conduit; raw question text never propagates \\
\rowcolor{bandExt} 2 & Day skeleton plus clue plan & Stage 1 outputs, plus 14-day windows $\to$ per-day theme, narrative arc, dominant mood, \texttt{planned\_events[]}, \texttt{clue\_plan} & reasoning LLM (\texttt{gpt-5-mini}, low effort) & 10 to 18 events per day; $\geq$50\% synthetic; anchors must appear \\
\rowcolor{rowAlt} 3 & Trajectory expansion & 5-day skeleton chunks $\to$ full anchor plus synthetic events with \texttt{description\_long} & reasoning LLM (\texttt{gpt-5-mini}, low effort) & verbatim preservation of anchor content; clue-kind taxonomy enforced \\
\rowcolor{bandQual} Time & Uniform representation & all events $\to$ 9 buckets plus per-day \texttt{seq} & rule-based & no calendar dates; no 12-h-clock times; \texttt{Day\{N\}} only \\
\rowcolor{rowAlt} Validator & Structural QA & expanded chunk $\to$ pass/fail & rule-based & exact equality of anchor content, time-bucket consistency, total-count match, privacy blacklist \\
\bottomrule
\end{tabularx}
\arrayrulecolor{black}
\end{table}

\paragraph{Why the \texttt{survey\_insight} bottleneck matters.}
A naive synthesizer that re-reads the 100 survey questions while writing the trajectory would inevitably restate them in prose. Stage 1 instead distills the survey into a four-part \texttt{profile\_persona.json} object whose only survey-derived field is \texttt{survey\_insight}: an 8 to 15 paragraph implicit-preference summary plus 5 to 12 cross-axis deep inferences and a domain-indexed view. Stages 2 to 3 receive only this distilled object, never the raw 100 questions. As a consequence, the trajectory cannot mechanically copy a question's wording, and survey answers can only enter via behavioural cues (a late dinner, a re-ordered errand, a sigh after a meeting).

\paragraph{Anchor regime vs.\ synthetic regime.}
At Stage 3, every event carries a provenance flag distinguishing anchor from synthetic. Anchor events reproduce the original log entry verbatim, retaining meal descriptions, mood annotations, proactive-help labels and rationales, photographs, and timing fields, while the model is permitted only to author a third-person narrative paraphrase of what the participant did. Synthetic events enjoy full content freedom but are conditioned on (a) the day's theme and narrative arc, (b) the participant's behavioural signature and persona texture, and (c) any planting plan entries whose target list intersects this event. Each planted clue carries a kind drawn from \{timing-detail, consequence, decision-frame, post-event-mood, social-context, app-choice, tiny-detail\} that constrains \emph{how} the answer is encoded behaviourally.

\paragraph{Time and date handling.}
We deliberately drop explicit \texttt{HH:MM} from synthetic events and use the nine fixed time buckets (\texttt{deep\_night}, \texttt{early\_morning}, \texttt{morning}, \texttt{noon}, \texttt{afternoon}, \texttt{dusk}, \texttt{evening}, \texttt{night}, \texttt{late\_night}) plus a per-day \texttt{seq} integer; anchor events retain whatever timing fields the original log carried (\texttt{wakeTime}, \texttt{sleepTime}, \texttt{leaveHomeTime}, \texttt{returnHomeTime}). This decision both enforces privacy and discourages models from over-fitting to spurious clock-time correlations.

\paragraph{Validator rules.}
A rule-based post-hoc validator enforces (i) exact equality between every anchor \texttt{payload} and the original log entry; (ii) consistency between each event's \texttt{time\_bucket} and the skeleton's \texttt{planned\_events}; (iii) total-event-count match between Stage 2 (skeleton) and Stage 3 (expansion); (iv) a privacy blacklist that rejects calendar dates, weekday names, real names, and 12-hour-clock times.

\subsection{Composition of the released artefact}
\label{sec:app-composition}

\begin{figure}[!htbp]
\centering
\includegraphics[width=0.80\linewidth]{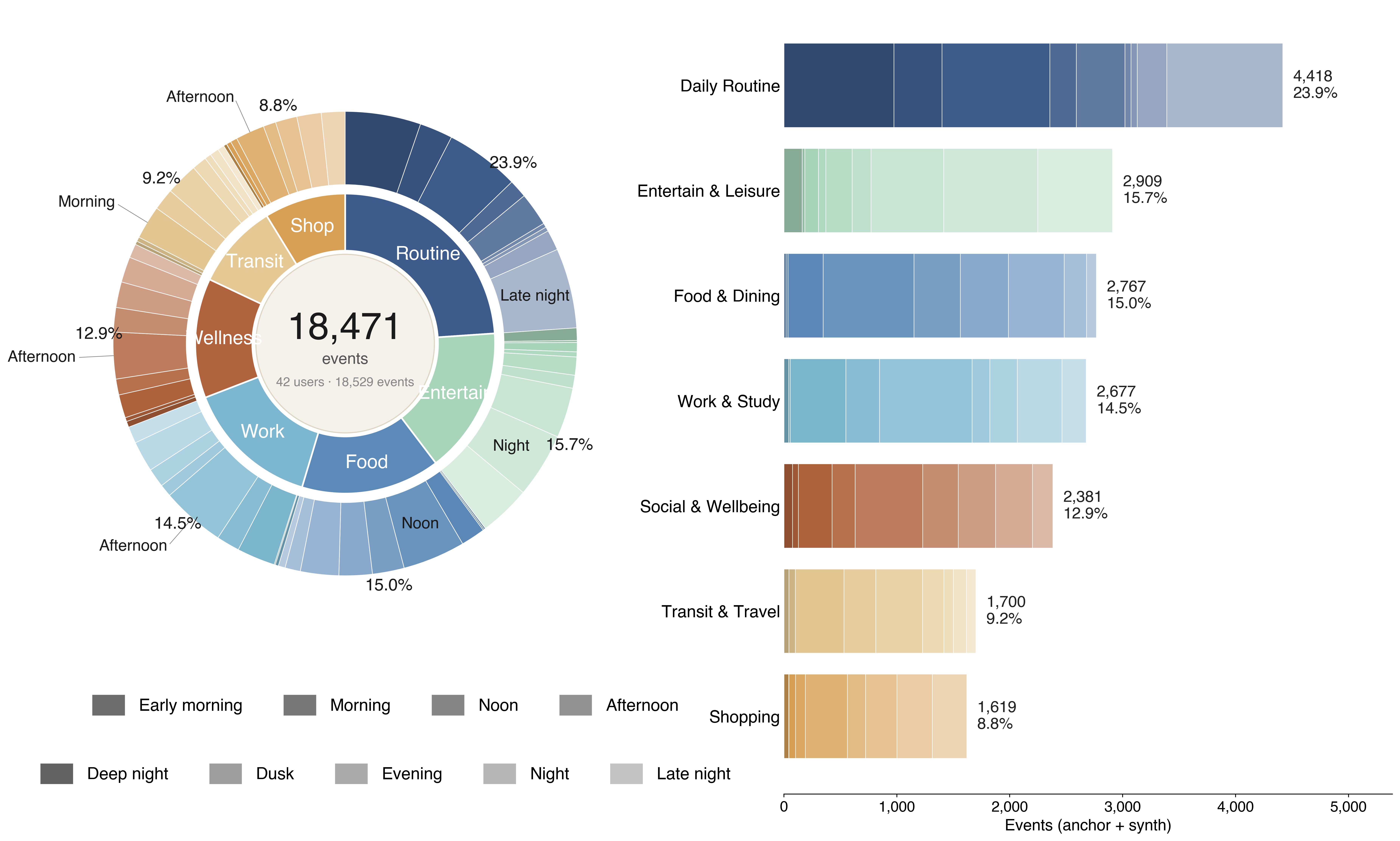}
\caption{\textbf{Composition of \textsc{Jarvis-Bench} by event count (18,471 events across 42 users).} \emph{Left}: nested donut showing the 7-lifestyle-domain breakdown (inner ring) and the 9-bucket time-of-day distribution within each domain (outer ring). \emph{Right}: per-domain stacked bars over the 9 time buckets, with absolute event counts and percentages. Daily Routine is the largest bucket (23.9\%), followed by Entertainment \& Leisure (15.7\%), Food \& Dining (15.0\%), Work \& Study (14.5\%), Social \& Wellbeing (12.9\%), Transit \& Travel (9.2\%), and Shopping (8.8\%). Within every domain the nine time buckets are populated, with Daily Routine concentrated in deep-night and early-morning slots, while Work, Food, and Entertainment span the daytime hours.}
\label{fig:composition-events}
\end{figure}

\begin{figure}[!htbp]
\centering
\includegraphics[width=0.80\linewidth]{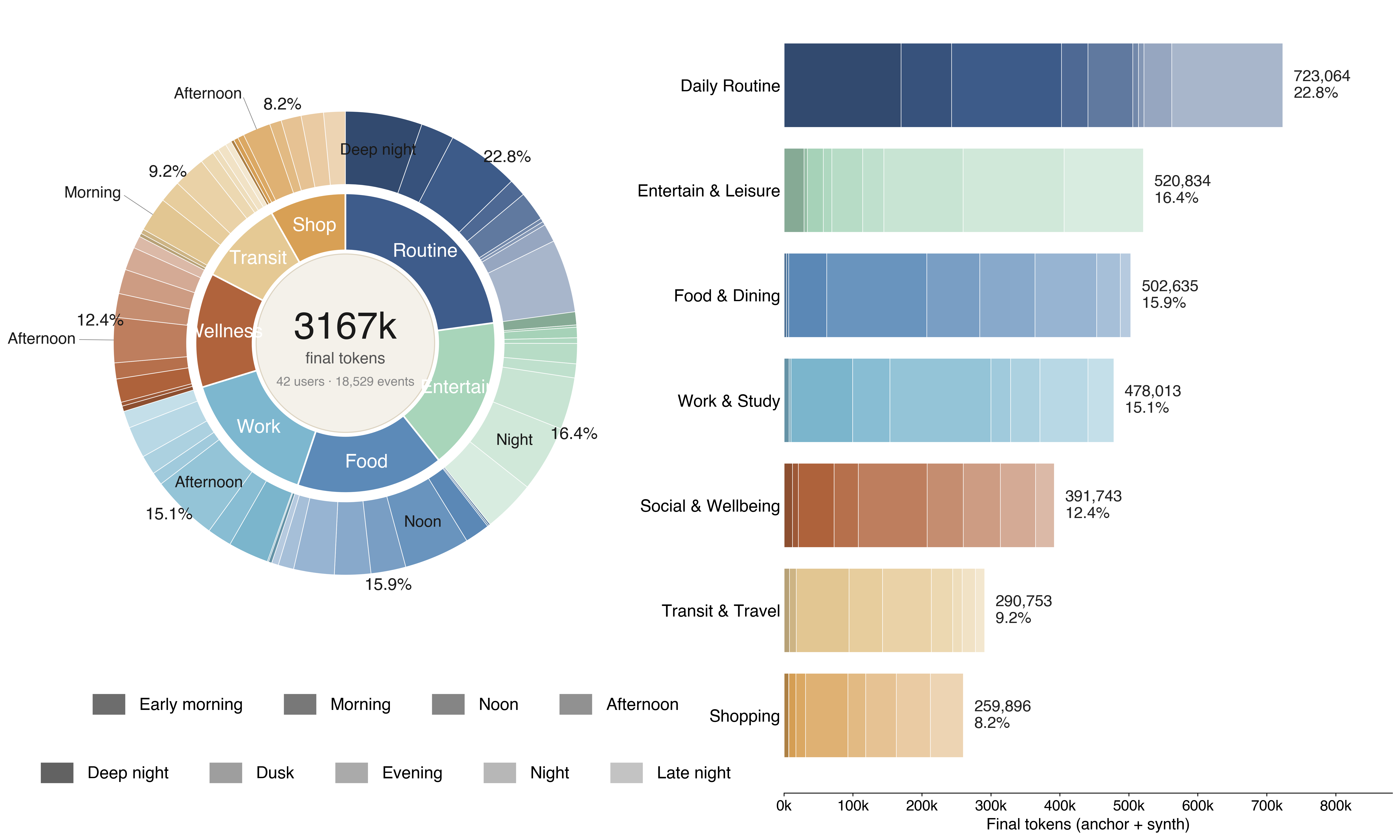}
\caption{\textbf{Composition of \textsc{Jarvis-Bench} by token volume ($\sim$3.17M final tokens across 42 users).} Same layout as \cref{fig:composition-events}, but cells are weighted by final-token count (anchor + synth) rather than event count. Token-weighted shares (Daily Routine 22.8\%, Entertainment \& Leisure 16.4\%, Food \& Dining 15.9\%, Work \& Study 15.1\%, Social \& Wellbeing 12.4\%, Transit \& Travel 9.2\%, Shopping 8.2\%) closely track the event-count shares in \cref{fig:composition-events}, indicating that no lifestyle domain is over- or under-described relative to its event frequency. The time-of-day spread within each domain bar mirrors the event-level distribution.}
\label{fig:composition-tokens}
\end{figure}

\begin{figure}[!htbp]
\centering
\includegraphics[width=0.96\linewidth]{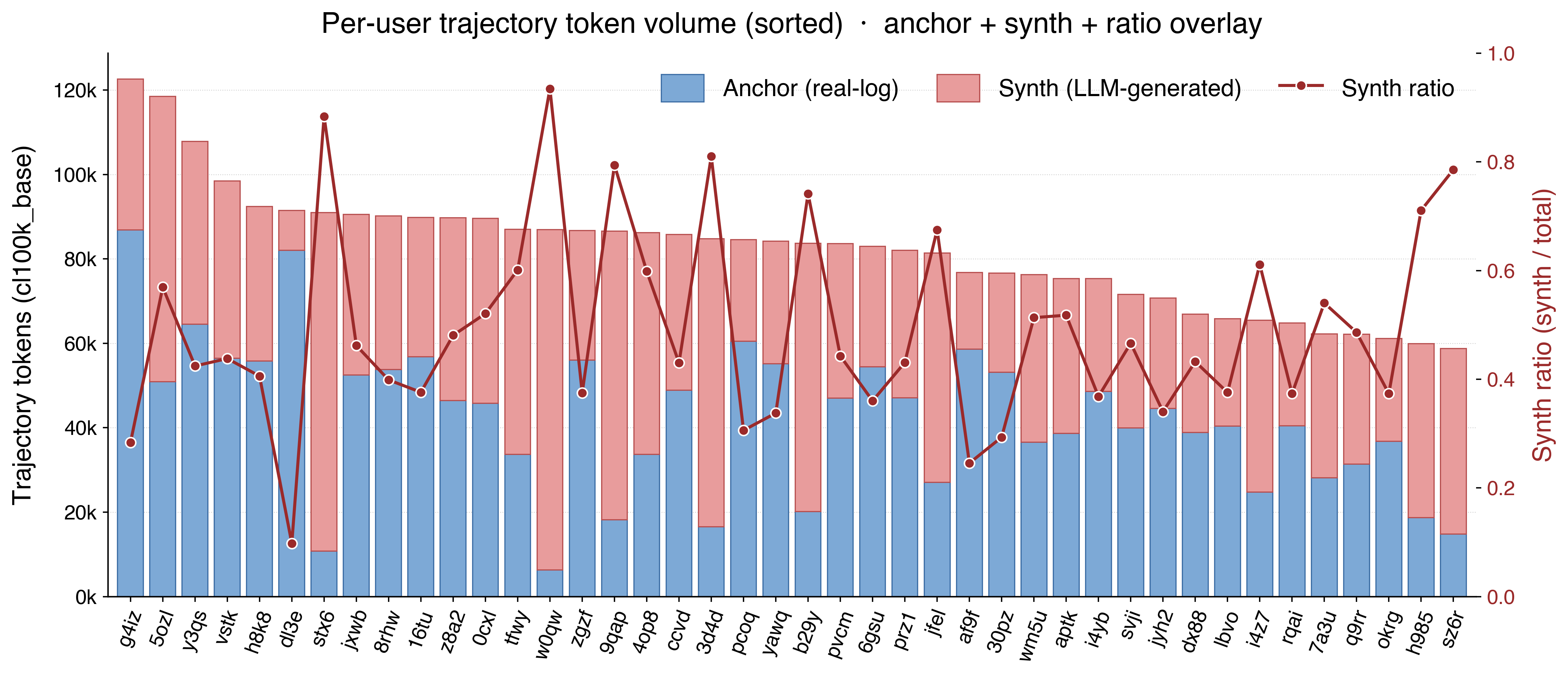}
\caption{\textbf{Per-user trajectory token volume (anchor + synth) with synth-ratio overlay.} Each bar is one of the 42 retained participants, sorted by total token count (descending). Anchor tokens (\emph{blue}, real-log) and synthetic tokens (\emph{pink}, LLM-generated) are stacked; the red line plots the per-user synth ratio (synth $/$ total) on the right axis. Most participants fall in the 60--100k token range with a synth ratio between 0.3 and 0.6; a small number of users with sparser self-tracking exhibit higher synth ratios ($\sim$0.7--0.8) because the densification step contributed proportionally more to fill the trajectory. The composition is \emph{anchor-completion, not anchor-replacement}: every real anchor token remains untouched, and synthetic tokens surround it.}
\label{fig:composition-per-user}
\end{figure}

We decompose the released artefact along three views: by event count and time-of-day spread (\cref{fig:composition-events}), by token volume and time-of-day spread (\cref{fig:composition-tokens}), and at the per-user level (\cref{fig:composition-per-user}). Across the 42 retained participants, the released set comprises \textbf{18,471 evaluated events} ($\sim$18,529 raw events before deduplication) totalling \textbf{$\sim$3.17 million tokens}. The lifestyle-domain mix is broadly balanced across both event and token views, and time-of-day coverage is wide within every domain, confirming that the synthesis layer succeeds in filling the deep-night, morning, work, and commuting surfaces that voluntary self-tracking under-reports.

\FloatBarrier
\subsection{Closing the loop: quality assurance and longitudinal participant feedback}
\label{sec:app-loop}

A trajectory is admitted to the benchmark only after passing three checks:

\begin{enumerate}[leftmargin=*, nosep]
\item \textbf{Structural validator (automatic).} A rule-based check enforces (i) exact equality between every anchor \texttt{payload} and the original log; (ii) consistency between each event's \texttt{time\_bucket} and the day skeleton; (iii) total-event-count match between skeleton and expansion; (iv) the privacy blacklist.
\item \textbf{Per-user full-trajectory annotation pass.} An internal reviewer reads each released trajectory and flags structural anomalies (mood inconsistency, narrative jumps, anchor and synthetic incongruities); flagged users are sent back through the expansion stage with a corrective prompt and re-validated.
\item \textbf{Participant fidelity confirmation.} For each participant, we send back a fully assembled trajectory with their original anchor entries highlighted and a randomly sampled subset of synthetic events foregrounded for review, and ask whether the assembled trajectory still feels like a plausible stretch of their own life. Negative responses trigger regeneration with stricter persona conditioning, and a participant is admitted to the released set only upon explicit confirmation. This step is what makes the qualified set genuinely real-user grounded rather than merely real-anchor-seeded.
\end{enumerate}

All numbers reported in the rest of the paper are computed on the 42 retained users.

\clearpage
\section{Participant Diversity}
\label{app:participant-diversity}
\begin{figure}[!htbp]
\centering
\begin{minipage}[t]{0.32\linewidth}
\centering
\includegraphics[width=\linewidth]{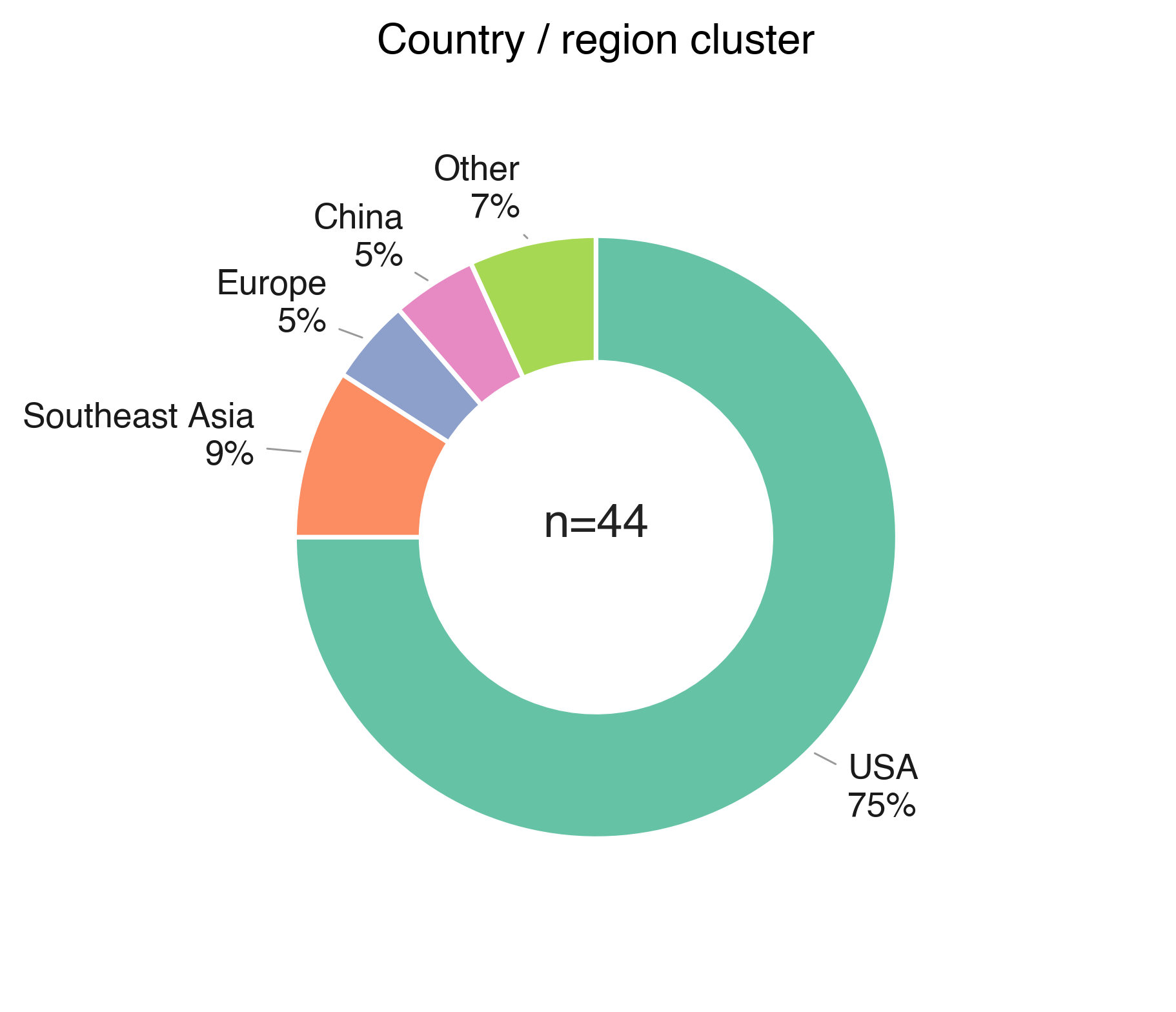}
\end{minipage}\hfill
\begin{minipage}[t]{0.32\linewidth}
\centering
\includegraphics[width=\linewidth]{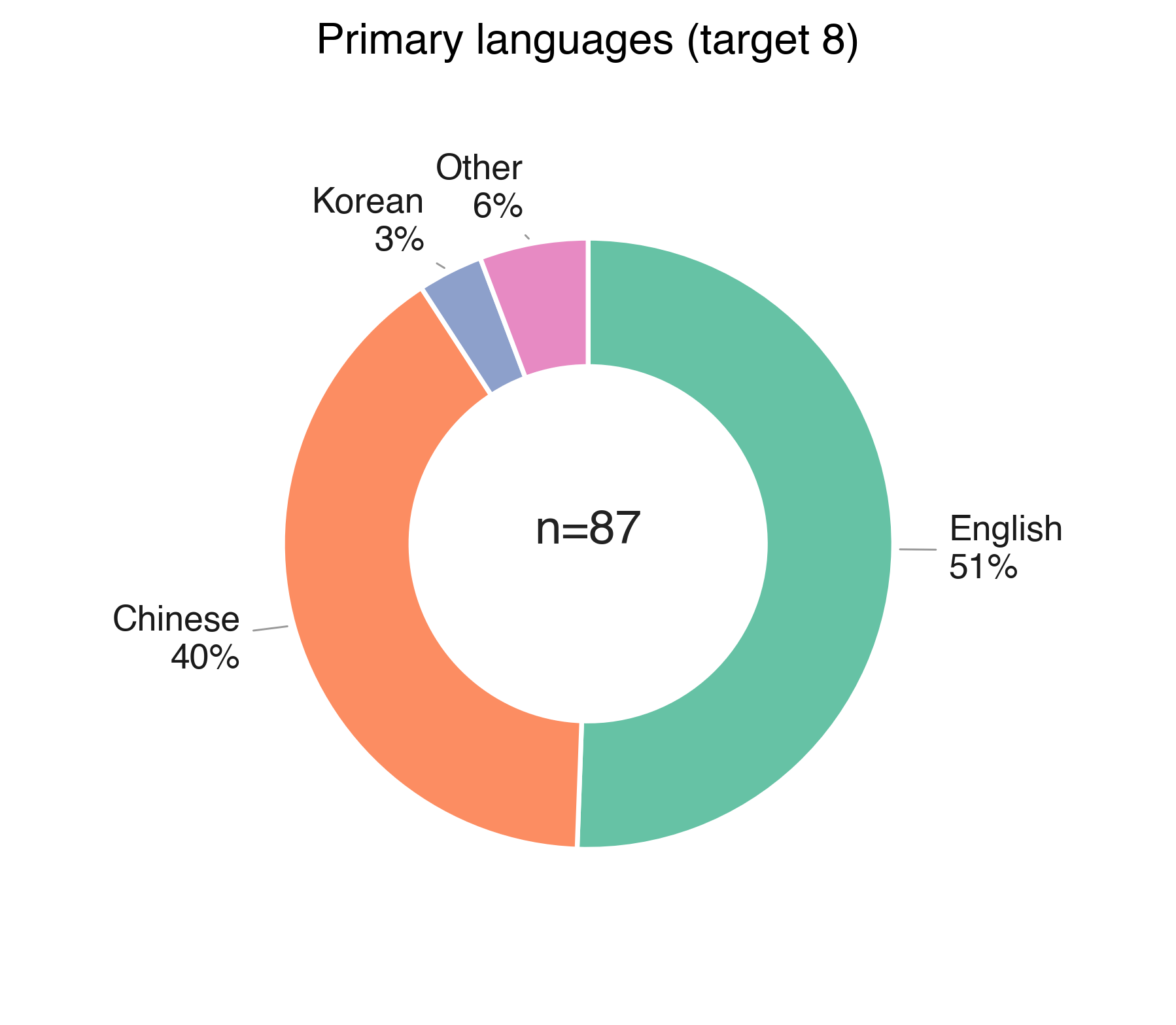}
\end{minipage}\hfill
\begin{minipage}[t]{0.32\linewidth}
\centering
\includegraphics[width=\linewidth]{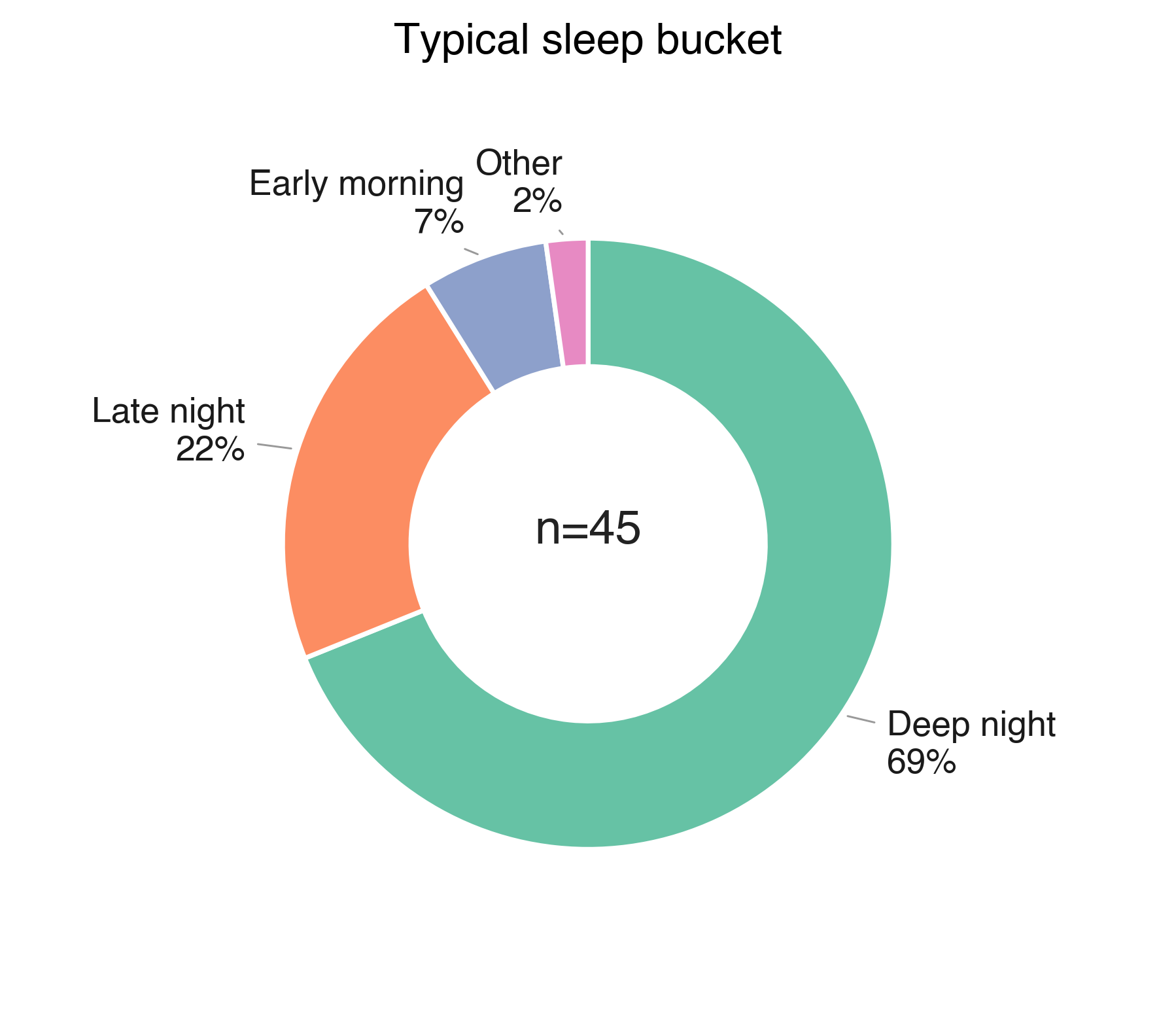}
\end{minipage}

\vspace{0.6em}

\begin{minipage}[t]{0.32\linewidth}
\centering
\includegraphics[width=\linewidth]{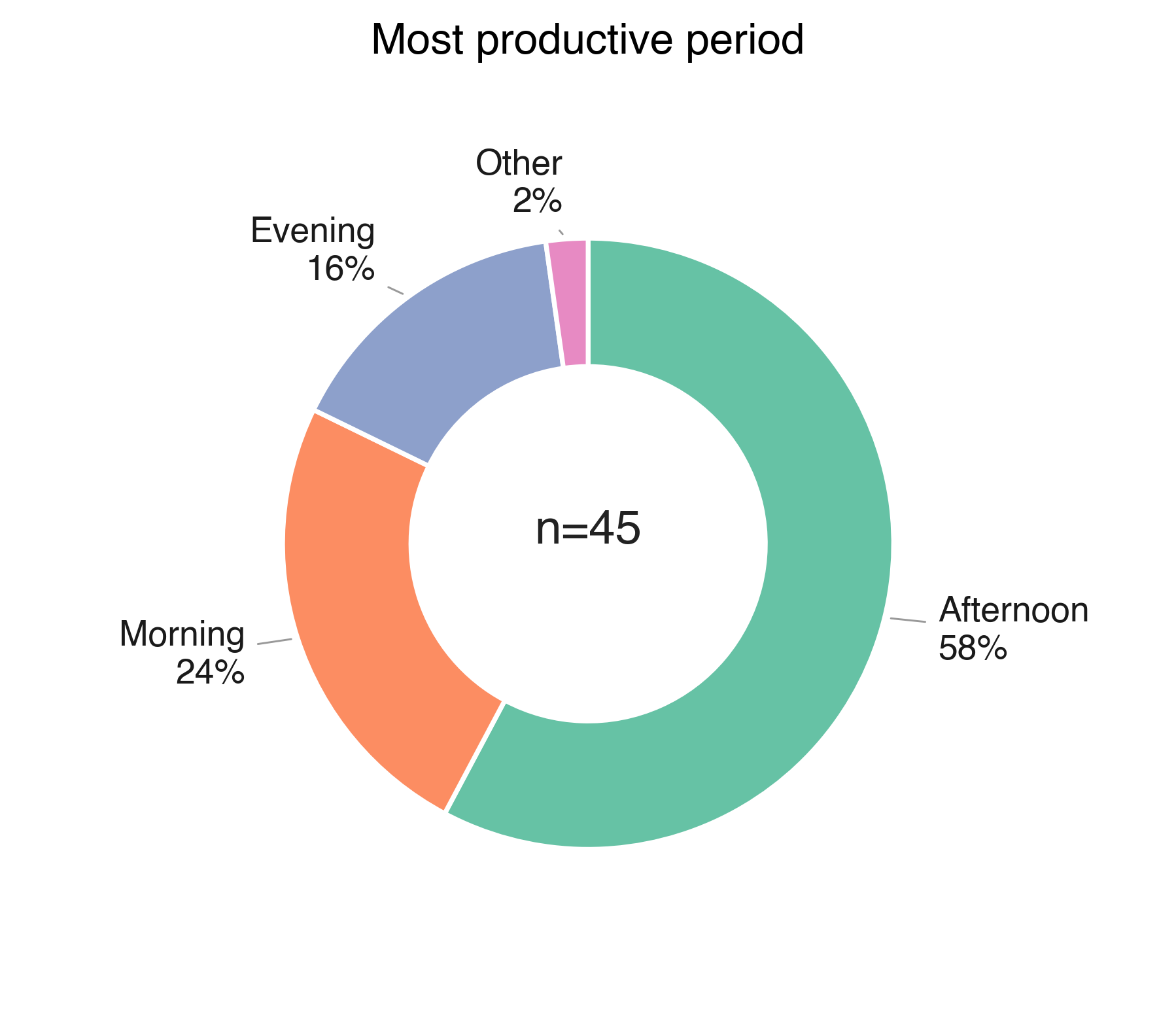}
\end{minipage}\hfill
\begin{minipage}[t]{0.32\linewidth}
\centering
\includegraphics[width=\linewidth]{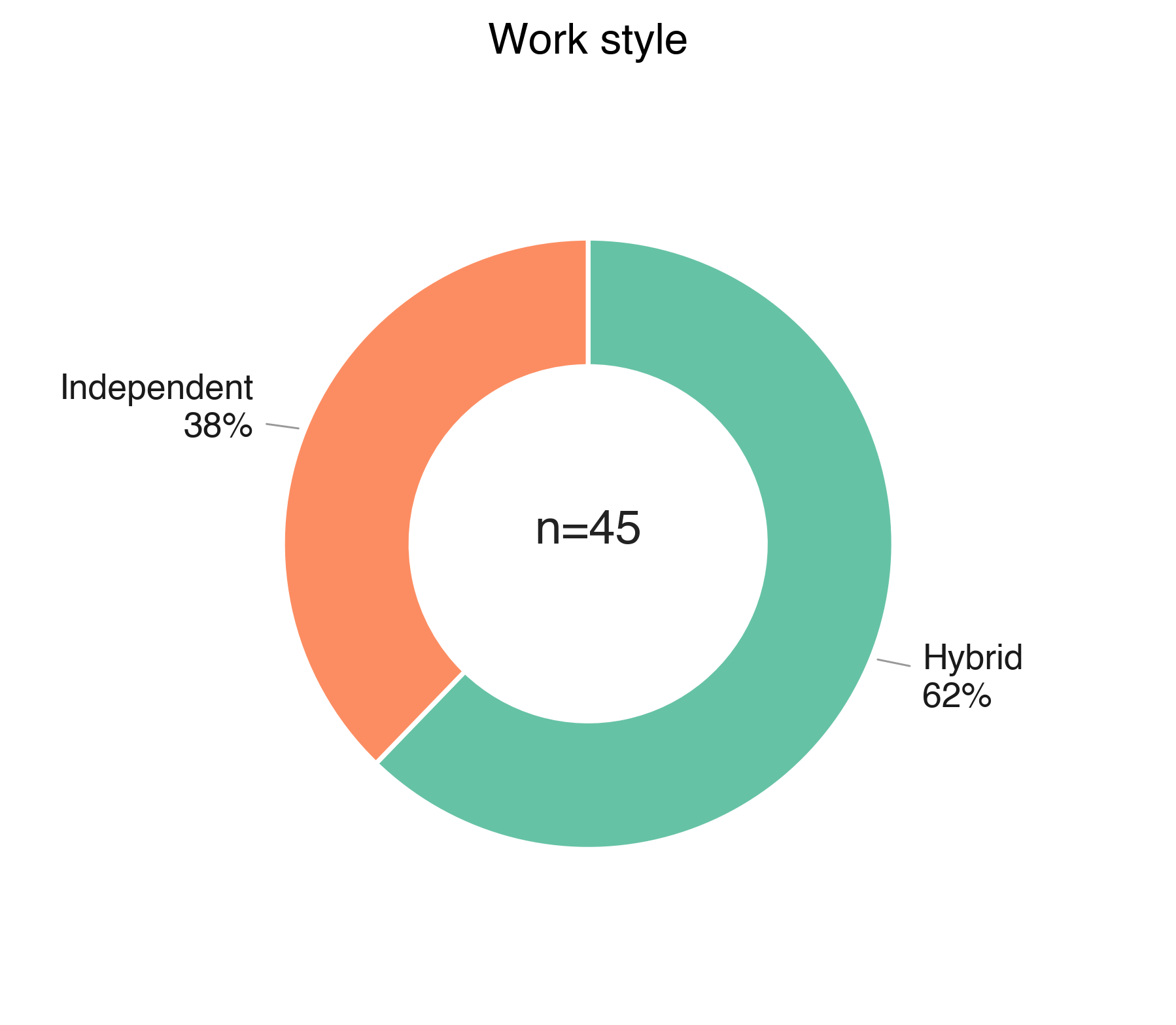}
\end{minipage}\hfill
\begin{minipage}[t]{0.32\linewidth}
\centering
\includegraphics[width=\linewidth]{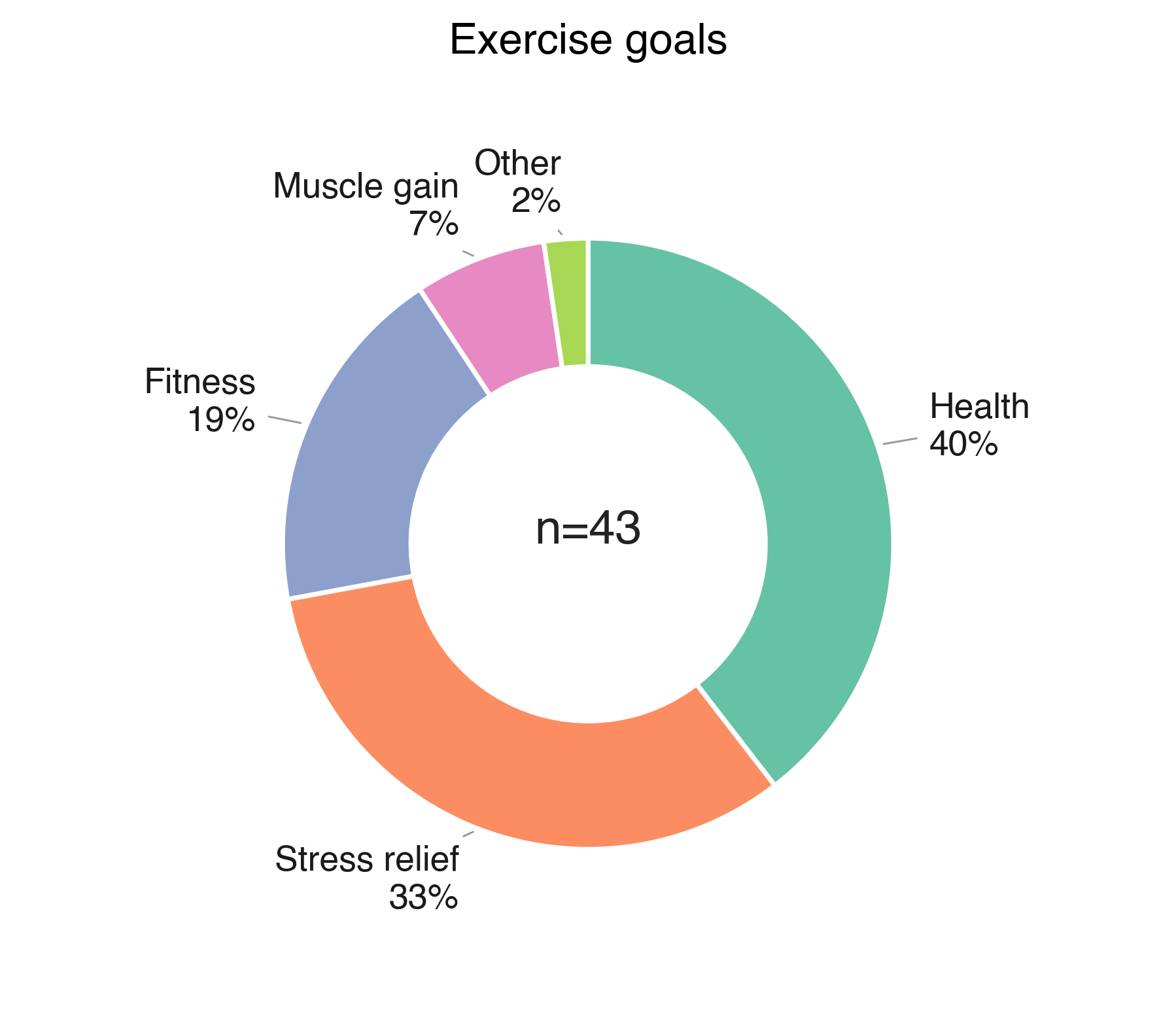}
\end{minipage}

\caption{\textbf{Participant diversity across six self-reported axes.} Top row: geographic region (USA 75\%, Southeast Asia 9\%, Europe 5\%, China 5\%, Other 7\%), primary language (English 51\%, Chinese 40\%, Korean 3\%, Other 6\%), and typical sleep bucket (Deep night 69\%, Late night 22\%, Early morning 7\%, Other 2\%). Bottom row: most productive period (Afternoon 58\%, Morning 24\%, Evening 16\%, Other 2\%), work style (Hybrid 62\%, Independent 38\%), and exercise goals (Health 40\%, Stress relief 33\%, Fitness 19\%, Muscle gain 7\%, Other 2\%). Sample sizes vary by question because some axes were optional and a small number of participants selected multiple categories that contribute to a single canonical bucket.}
\label{fig:participant-diversity}
\end{figure}

This appendix summarises the demographic and lifestyle diversity of the participant pool that contributes to \textsc{Jarvis-Bench}. \Cref{fig:participant-diversity} presents six donut breakdowns spanning geographic, linguistic, chronotype, work-mode, and lifestyle axes. The pool is geographically anchored in the United States (75\%) but augmented with Southeast Asia (9\%), Europe (5\%), and China (5\%); the language mix is split roughly evenly between English (51\%) and Chinese (40\%) primary speakers, with Korean and other languages making up the remainder. Chronotype evidence shows a late-shifted pool (69\% of participants typically sleep in the deep-night bucket and 22\% in the late-night bucket), while self-reported peak productivity concentrates in the afternoon (58\%). The work-style split is dominated by hybrid arrangements (62\%), and reported exercise motivations span health (40\%), stress relief (33\%), fitness (19\%), and muscle gain (7\%). Together, these axes confirm that the qualified participants span a broad slice of lifestyle archetypes rather than collapsing onto a single demographic mode.

\clearpage
\section{Implicit Preference Reasoning: Full Results}
\label{app:ipq}
This appendix collects all results and stratifications for the Implicit Preference Reasoning task. The main body keeps only \cref{fig:main} (the user-memory ladder for all 54 LLMs) and four findings; everything else is here.

\subsection{User-memory ladder}
\label{sec:app-ipq-conditions}

The five conditions form a single user-memory ladder, which Implicit Preference Reasoning uses in full; \cref{tab:conditions} maps each display name to its underlying configuration. The display names are \emph{No Memory}, \emph{Short Trajectory}, \emph{Long Trajectory}, \emph{Persona Memory}, and \emph{Lifelong Memory}. The released code and CSV files use the condition codes \texttt{C0\_blank} through \texttt{C4\_skyline}; the paper uses the display names. Lifelong Memory (C4) is an oracle upper bound: the model receives the same distilled survey-insight summary that conditioned the synthesis pipeline. Note that Lifelong Memory does \emph{not} reveal raw question text or answer letters; it is the model's view of an ideally-prepared personalization context.

\begin{table}[h]
\centering
\caption{\textbf{User-memory ladder (No Memory to Lifelong Memory).} Implicit Preference Reasoning is evaluated under all five conditions, while Proactive Help and Emotion-Aware Modeling use the history preceding the target. The display name is what the paper uses; the code is the internal CSV identifier. The Cross-Domain Transfer task additionally uses S1 / S2 ablations derived from Long Trajectory / No Memory.}
\label{tab:conditions}
\small
\arrayrulecolor{tabHeader}
\renewcommand{\arraystretch}{1.18}
\begin{tabular}{@{}llcccp{2.6cm}@{}}
\toprule
\rowcolor{tabHeader}
\hdrcell{Display name} & \hdrcell{Code} & \hdrcell{Trajectory} & \hdrcell{Persona} & \hdrcell{Survey insight} & \hdrcell{Purpose} \\
\midrule
\textbf{No Memory}      & C0\_blank         & \no  & \no  & \no  & no-context baseline \\
\rowcolor{rowAlt} \textbf{Short Trajectory} & C1\_traj14d  & first 14 days & \no  & \no  & sparse-context probe \\
\textbf{Long Trajectory}     & C2\_full     & all days      & \no  & \no  & dense-context probe \\
\rowcolor{rowAlt} \textbf{Persona Memory}   & C3\_full\_persona & all days  & \yes & \no  & adds explicit profile \\
\rowcolor{bandHigh} \textbf{Lifelong Memory} & C4\_skyline & all days      & \yes & \yes & oracle ceiling \\
\bottomrule
\end{tabular}
\arrayrulecolor{black}
\end{table}

\subsection{Per-domain accuracy across the memory ladder}
\label{sec:app-ipq-perdomain}

\begin{figure}[!htbp]
\centering
\includegraphics[width=\linewidth]{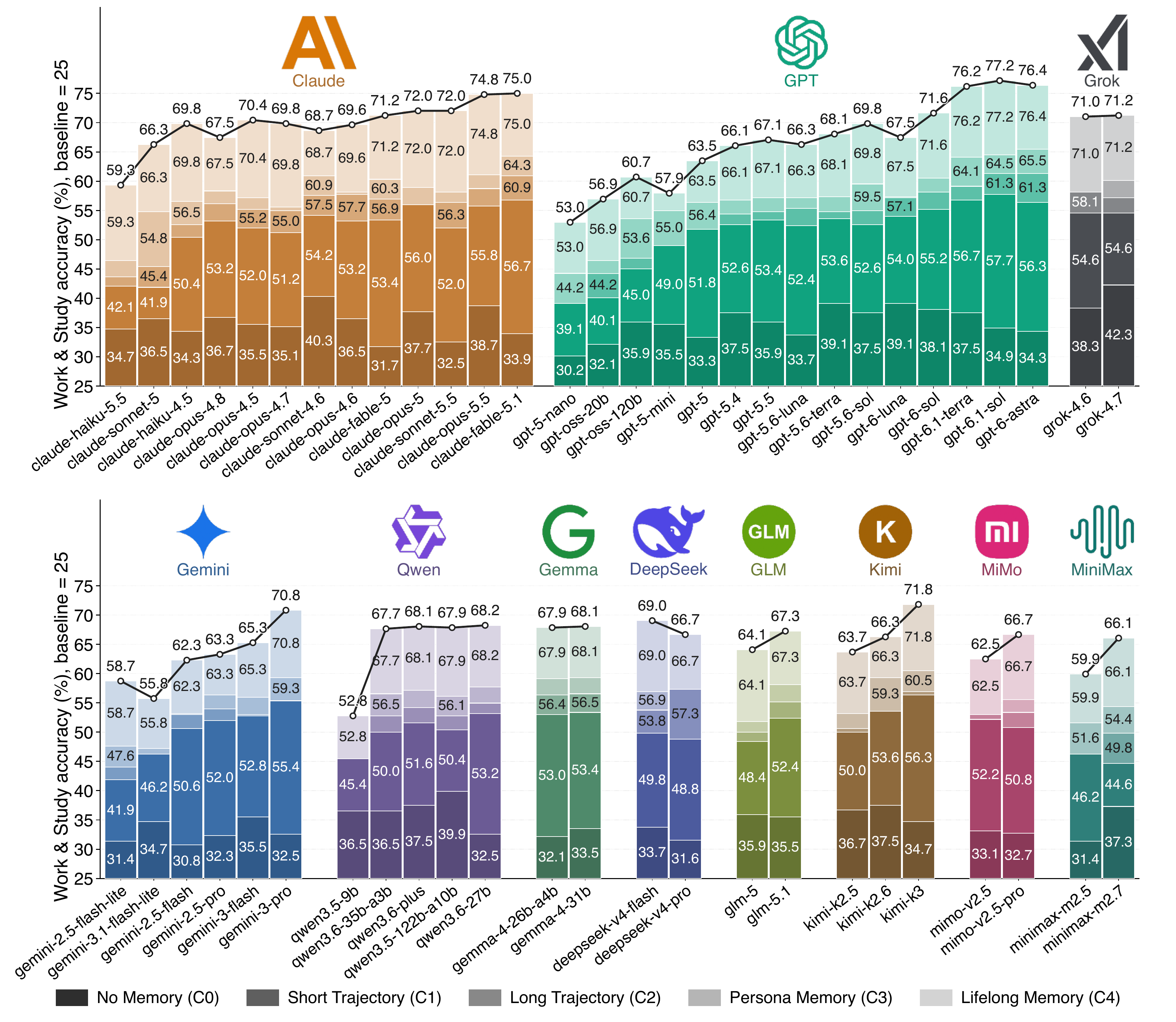}
\caption{\textbf{Per-domain accuracy on \emph{Work \& Study} questions across the user-memory ladder, all 54 LLMs.} Same layout as \cref{fig:main}. Random baseline = 25\%. GPT-6.1-Sol reaches the highest accuracy on this domain (77.2\% at C4), and Gemini-3-Pro leads its family at 70.8\%; mid-size models (Qwen3.6-27B, Gemma-4-31B, Claude-Sonnet-4.6) cluster in the high-60\% range; the GPT-5 family lands lower, with GPT-5-Nano below 60\% even at C4. Work \& Study is broadly accessible: most families exceed 60\% at C4, suggesting that work-related preference cues are well-represented in the trajectory.}
\label{fig:domain-workstudy}
\end{figure}

\begin{figure}[!htbp]
\centering
\includegraphics[width=\linewidth]{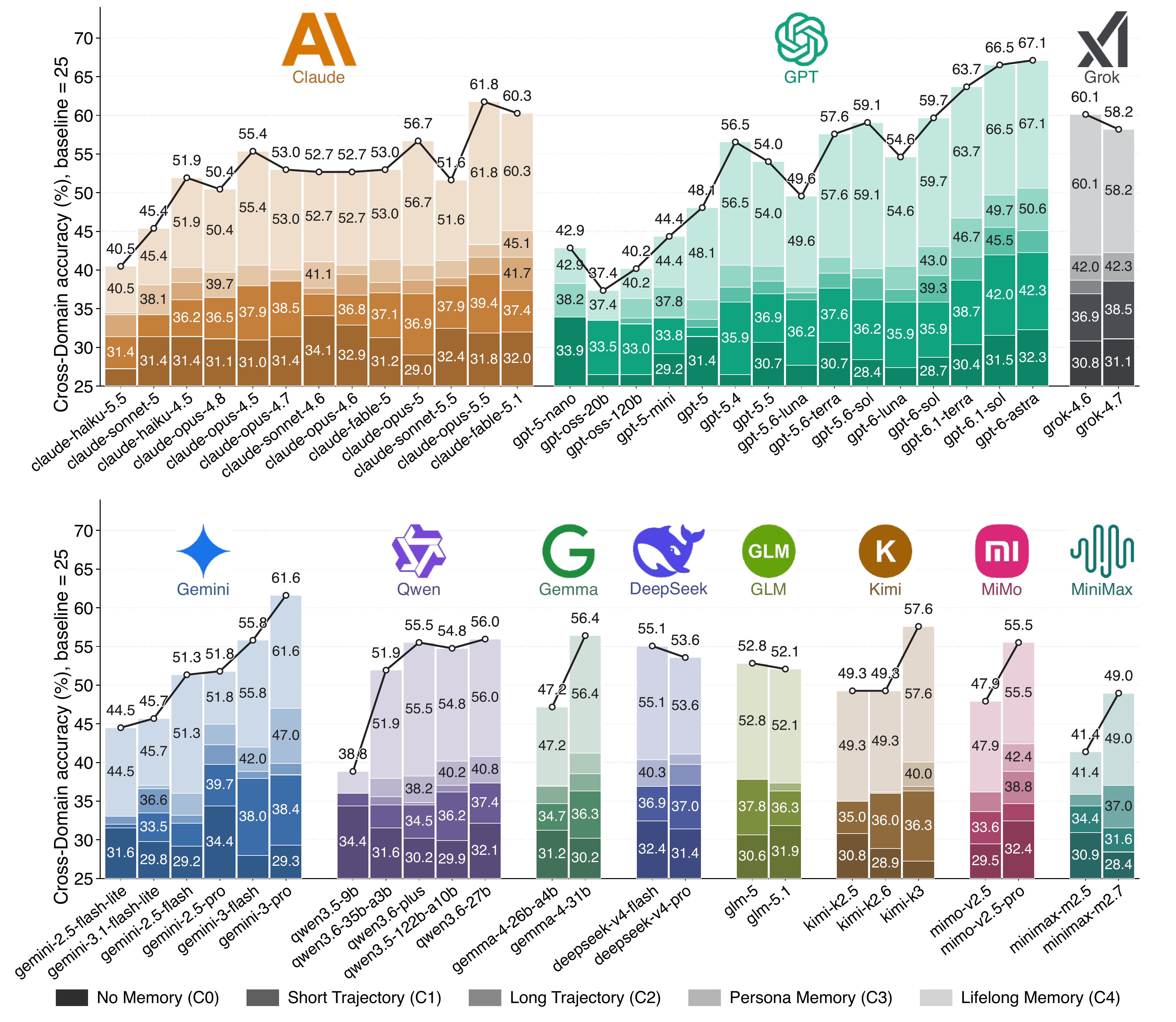}
\caption{\textbf{Per-domain accuracy on \emph{Cross-Domain} questions across the user-memory ladder, all 54 LLMs.} Same layout as \cref{fig:domain-workstudy}. Cross-Domain is the hardest lifestyle bucket: even at Lifelong Memory (C4) the strongest model (GPT-6-Astra) reaches only 67.1\%, Gemini-3-Pro leads its family at 61.6\%, and most open-weight families top out in the low- to mid-50\% range; the GPT-5 family ceilings below 50\%, and the smallest models (Qwen3.5-9B, GPT-5-Nano, and the two Gemini Flash-Lite models) sit between 39\% and 46\%. Because Cross-Domain questions require integrating preference evidence \emph{across} multiple lifestyle areas (food $\times$ leisure, work $\times$ social, etc.), their persistent difficulty even under oracle context is consistent with the Cross-Domain Transfer finding (\cref{sec:results-transfer}) that current models fail to compose preferences across lifestyle boundaries.}
\label{fig:domain-crossdomain}
\end{figure}

We report per-domain breakdowns across the full five-condition memory ladder for two contrasting lifestyle domains: \emph{Work \& Study} (\cref{fig:domain-workstudy}) as a representative everyday domain that benefits steadily from richer memory, and \emph{Cross-Domain} (\cref{fig:domain-crossdomain}) as the residual-difficulty bucket whose questions require integrating evidence across multiple lifestyle areas. The same plot type is generated for every lifestyle domain in the released CSV (\texttt{by\_domain\_\_average.csv}); we display these two as illustrative endpoints, since the remaining domains interpolate between them. Across both plots, every family's accuracy increases monotonically from C0 (No Memory, dark) to C4 (Lifelong Memory, light), and the spread between C0 and C4 is the per-domain analogue of the headline finding that trajectory density and oracle context together account for most of the personalization lift.

\FloatBarrier
\subsection{Stratifications by clue difficulty, clue type, and planted vs.\ unplanted}
\label{sec:app-ipq-clue}

\paragraph{By clue difficulty.}
We split the planted questions (about 20 per user) into \texttt{hard} ($\sim$2.5 per user) and \texttt{very\_hard} ($\sim$17.6 per user) groups. At No Memory both groups sit at $\sim$0.34 (chance plus lexical priors). At Lifelong Memory the \texttt{hard} group reaches $\sim$0.74 and the \texttt{very\_hard} group reaches $\sim$0.66; the 8-point gap is exactly the part of the survey we designed to require multi-hop integration. Short Trajectory already lifts \texttt{very\_hard} by $\sim$22 points over No Memory, indicating that even a 14-day window plants enough behavioural cues for many ``very hard'' answers.

\paragraph{By clue type.}
Stratifying by \texttt{clue\_type}, \texttt{multi\_hop} and \texttt{cross\_domain} clues benefit most from the No Memory to Lifelong Memory transition; \texttt{tiny\_detail} clues remain hardest at Lifelong Memory (typical accuracy $\sim$0.50 vs.\ $\sim$0.68 for \texttt{multi\_hop}), confirming that \emph{thinly-encoded} preferences are the residual difficulty even when the model has full context.

\paragraph{Planted vs.\ unplanted.}
The planted-vs-unplanted gap is small at No Memory ($\sim$0.36 vs.\ $\sim$0.34) but widens to $\sim$5 to 7 points at Lifelong Memory. Models reliably exploit the planted evidence when it is present, but the trajectory's incidental coverage helps unplanted questions less.

\subsection{Full accuracy matrix}
\label{sec:app-ipq-fullmatrix}
\label{sec:app-ipq-new}

\Cref{tab:matrix-a,tab:matrix-b} report Implicit Preference Reasoning accuracy for all 54 LLMs on the user-memory ladder, grouped by family as in \cref{fig:main}. For a few DeepSeek, GLM, Kimi-K2, MiMo, and MiniMax checkpoints some users are missing in some conditions (logging or vendor outage); the per-condition $n$ is preserved in the released CSV. Random baseline = 0.25. GPT-6-Astra holds the highest Lifelong-Memory (0.732) and Long-Trajectory (0.550) accuracy. User-level bootstrap 95\% confidence intervals for the Bedrock-served models have half-widths of 1.4 to 2.4 points at C2 and C4, so gaps of one or two points between models are within sampling noise.

\begin{table}[h]
\centering
\caption{\textbf{Implicit Preference Reasoning accuracy across the user-memory ladder: Claude, GPT, and Grok families.} Columns: \emph{No Memory} (C0), \emph{Short Trajectory} (C1, 14d), \emph{Long Trajectory} (C2, full), \emph{Persona Memory} (C3, full + persona), \emph{Lifelong Memory} (C4, oracle). \textsc{Avg} = mean of the five conditions. Lifelong-Memory and \textsc{Avg} best per family in bold; the best Long-Trajectory and Lifelong-Memory values across all 54 models shaded.}
\label{tab:matrix-a}
\small
\arrayrulecolor{tabHeader}
\setlength{\tabcolsep}{4pt}
\renewcommand{\arraystretch}{1.10}
\begin{tabular}{@{}l|cccccc@{}}
\toprule
\rowcolor{tabHeader}
\hdrcell{Model} & \hdrcell{\makecell{No\\Memory\\\scriptsize(C0)}} & \hdrcell{\makecell{Short\\Trajectory\\\scriptsize(C1)}} & \hdrcell{\makecell{Long\\Trajectory\\\scriptsize(C2)}} & \hdrcell{\makecell{Persona\\Memory\\\scriptsize(C3)}} & \hdrcell{\makecell{Lifelong\\Memory\\\scriptsize(C4)}} & \hdrcell{Avg} \\
\midrule
\rowcolor{bandClaude}\multicolumn{7}{@{}l@{}}{\,\textbf{\textcolor{tabHeader}{Claude family}}} \\
claude-fable-5.1               & 0.354 & 0.500 & 0.540 & 0.572 & \textbf{0.700} & \textbf{0.533} \\
\rowcolor{rowAlt} claude-opus-5.5                & 0.366 & 0.494 & 0.527 & 0.557 & 0.695 & 0.528 \\
claude-opus-5                  & 0.366 & 0.480 & 0.494 & 0.525 & 0.651 & 0.503 \\
\rowcolor{rowAlt} claude-fable-5                 & 0.359 & 0.480 & 0.499 & 0.533 & 0.639 & 0.502 \\
claude-sonnet-5.5              & 0.355 & 0.466 & 0.497 & 0.537 & 0.651 & 0.501 \\
\rowcolor{rowAlt} claude-opus-4.6                & 0.361 & 0.474 & 0.500 & 0.521 & 0.631 & 0.497 \\
claude-sonnet-4.6              & 0.362 & 0.466 & 0.493 & 0.522 & 0.631 & 0.495 \\
\rowcolor{rowAlt} claude-opus-4.5                & 0.348 & 0.464 & 0.478 & 0.520 & 0.645 & 0.491 \\
claude-opus-4.7                & 0.349 & 0.463 & 0.481 & 0.511 & 0.622 & 0.485 \\
\rowcolor{rowAlt} claude-haiku-4.5               & 0.330 & 0.449 & 0.466 & 0.511 & 0.625 & 0.476 \\
claude-opus-4.8                & 0.355 & 0.455 & 0.470 & 0.503 & 0.599 & 0.476 \\
\rowcolor{rowAlt} claude-sonnet-5                & 0.329 & 0.388 & 0.400 & 0.467 & 0.572 & 0.431 \\
claude-haiku-5.5               & 0.325 & 0.380 & 0.392 & 0.417 & 0.500 & 0.403 \\
\midrule
\rowcolor{bandGPT}\multicolumn{7}{@{}l@{}}{\,\textbf{\textcolor{tabHeader}{GPT family}}} \\
gpt-6-astra                    & 0.350 & 0.512 & \cellcolor{bandHigh}\textbf{0.550} & 0.589 & \cellcolor{bandHigh}\textbf{0.732} & \textbf{0.547} \\
\rowcolor{rowAlt} gpt-6.1-sol                    & 0.347 & 0.508 & 0.547 & 0.586 & 0.729 & 0.543 \\
gpt-6.1-terra                  & 0.343 & 0.492 & 0.524 & 0.569 & 0.712 & 0.528 \\
\rowcolor{rowAlt} gpt-6-luna                     & 0.346 & 0.479 & 0.508 & 0.537 & 0.645 & 0.503 \\
gpt-6-sol                      & 0.345 & 0.467 & 0.497 & 0.537 & 0.670 & 0.503 \\
\rowcolor{rowAlt} gpt-5.6-sol                    & 0.350 & 0.470 & 0.494 & 0.529 & 0.657 & 0.500 \\
gpt-5.6-terra                  & 0.353 & 0.476 & 0.498 & 0.524 & 0.646 & 0.499 \\
\rowcolor{rowAlt} gpt-5.5                        & 0.350 & 0.467 & 0.486 & 0.511 & 0.625 & 0.488 \\
gpt-5.6-luna                   & 0.342 & 0.469 & 0.488 & 0.517 & 0.614 & 0.486 \\
\rowcolor{rowAlt} gpt-5.4                        & 0.337 & 0.459 & 0.487 & 0.515 & 0.626 & 0.485 \\
gpt-5                          & 0.337 & 0.447 & 0.434 & 0.494 & 0.585 & 0.459 \\
\rowcolor{rowAlt} gpt-5-mini                     & 0.345 & 0.429 & 0.418 & 0.484 & 0.550 & 0.445 \\
gpt-oss-120b                   & 0.339 & 0.405 & 0.414 & 0.451 & 0.514 & 0.425 \\
\rowcolor{rowAlt} gpt-oss-20b                    & 0.325 & 0.400 & 0.405 & 0.449 & 0.510 & 0.418 \\
gpt-5-nano                     & 0.330 & 0.382 & 0.382 & 0.445 & 0.514 & 0.411 \\
\midrule
\rowcolor{bandExt}\multicolumn{7}{@{}l@{}}{\,\textbf{\textcolor{tabHeader}{Grok family}}} \\
grok-4.7                       & 0.364 & 0.488 & 0.508 & 0.546 & \textbf{0.678} & \textbf{0.517} \\
\rowcolor{rowAlt} grok-4.6                       & 0.349 & 0.480 & 0.508 & 0.545 & 0.676 & 0.512 \\
\bottomrule
\end{tabular}
\arrayrulecolor{black}
\end{table}

\begin{table}[h]
\centering
\caption{\textbf{Implicit Preference Reasoning accuracy across the user-memory ladder: Gemini, Qwen, Gemma, DeepSeek, GLM, Kimi, MiMo, and MiniMax families.} Same columns and marks as \cref{tab:matrix-a}.}
\label{tab:matrix-b}
\small
\arrayrulecolor{tabHeader}
\setlength{\tabcolsep}{4pt}
\renewcommand{\arraystretch}{1.10}
\begin{tabular}{@{}l|cccccc@{}}
\toprule
\rowcolor{tabHeader}
\hdrcell{Model} & \hdrcell{\makecell{No\\Memory\\\scriptsize(C0)}} & \hdrcell{\makecell{Short\\Trajectory\\\scriptsize(C1)}} & \hdrcell{\makecell{Long\\Trajectory\\\scriptsize(C2)}} & \hdrcell{\makecell{Persona\\Memory\\\scriptsize(C3)}} & \hdrcell{\makecell{Lifelong\\Memory\\\scriptsize(C4)}} & \hdrcell{Avg} \\
\midrule
\rowcolor{bandGemini}\multicolumn{7}{@{}l@{}}{\,\textbf{\textcolor{tabHeader}{Gemini family}}} \\
gemini-3-pro-preview           & 0.347 & 0.478 & 0.491 & 0.539 & \textbf{0.675} & \textbf{0.506} \\
\rowcolor{rowAlt} gemini-2.5-pro                 & 0.353 & 0.461 & 0.491 & 0.518 & 0.606 & 0.486 \\
gemini-3-flash-preview         & 0.329 & 0.463 & 0.480 & 0.506 & 0.618 & 0.479 \\
\rowcolor{rowAlt} gemini-2.5-flash               & 0.338 & 0.435 & 0.447 & 0.473 & 0.580 & 0.455 \\
gemini-2.5-flash-lite          & 0.340 & 0.397 & 0.413 & 0.438 & 0.534 & 0.425 \\
\rowcolor{rowAlt} gemini-3.1-flash-lite-preview  & 0.347 & 0.400 & 0.415 & 0.445 & 0.519 & 0.425 \\
\midrule
\rowcolor{bandQwen}\multicolumn{7}{@{}l@{}}{\,\textbf{\textcolor{tabHeader}{Qwen family}}} \\
qwen3.6-27b                    & 0.360 & 0.473 & 0.491 & 0.533 & \textbf{0.655} & \textbf{0.503} \\
\rowcolor{rowAlt} qwen3.5-122b-a10b              & 0.360 & 0.465 & 0.486 & 0.520 & 0.636 & 0.494 \\
qwen3.6-plus                   & 0.362 & 0.457 & 0.479 & 0.516 & 0.639 & 0.491 \\
\rowcolor{rowAlt} qwen3.6-35b-a3b                & 0.364 & 0.446 & 0.473 & 0.503 & 0.621 & 0.482 \\
qwen3.5-9b                     & 0.352 & 0.435 & 0.328 & 0.403 & 0.500 & 0.404 \\
\midrule
\rowcolor{bandGemma}\multicolumn{7}{@{}l@{}}{\,\textbf{\textcolor{tabHeader}{Gemma family}}} \\
gemma-4-31b                    & 0.348 & 0.471 & 0.494 & 0.522 & \textbf{0.652} & \textbf{0.497} \\
\rowcolor{rowAlt} gemma-4-26b-a4b                & 0.336 & 0.446 & 0.473 & 0.500 & 0.583 & 0.468 \\
\midrule
\rowcolor{bandExt}\multicolumn{7}{@{}l@{}}{\,\textbf{\textcolor{tabHeader}{DeepSeek family}}} \\
deepseek-v4-pro                & 0.352 & 0.471 & 0.501 & 0.515 & \textbf{0.631} & \textbf{0.494} \\
\rowcolor{rowAlt} deepseek-v4-flash              & 0.353 & 0.440 & 0.464 & 0.501 & 0.622 & 0.476 \\
\midrule
\rowcolor{bandExt}\multicolumn{7}{@{}l@{}}{\,\textbf{\textcolor{tabHeader}{GLM family}}} \\
glm-5.1                        & 0.362 & 0.467 & 0.483 & 0.510 & \textbf{0.621} & \textbf{0.489} \\
\rowcolor{rowAlt} glm-5                          & 0.347 & 0.443 & 0.455 & 0.472 & 0.594 & 0.462 \\
\midrule
\rowcolor{bandExt}\multicolumn{7}{@{}l@{}}{\,\textbf{\textcolor{tabHeader}{Kimi family}}} \\
kimi-k3                        & 0.345 & 0.480 & 0.504 & 0.541 & \textbf{0.665} & \textbf{0.507} \\
\rowcolor{rowAlt} kimi-k2.6                      & 0.364 & 0.463 & 0.478 & 0.505 & 0.606 & 0.483 \\
kimi-k2.5                      & 0.351 & 0.434 & 0.441 & 0.460 & 0.573 & 0.452 \\
\midrule
\rowcolor{bandExt}\multicolumn{7}{@{}l@{}}{\,\textbf{\textcolor{tabHeader}{MiMo family}}} \\
mimo-v2.5-pro                  & 0.345 & 0.463 & 0.485 & 0.522 & \textbf{0.636} & \textbf{0.490} \\
\rowcolor{rowAlt} mimo-v2.5                      & 0.334 & 0.441 & 0.450 & 0.478 & 0.583 & 0.457 \\
\midrule
\rowcolor{bandExt}\multicolumn{7}{@{}l@{}}{\,\textbf{\textcolor{tabHeader}{MiniMax family}}} \\
minimax-m2.7                   & 0.341 & 0.402 & 0.447 & 0.479 & \textbf{0.593} & \textbf{0.452} \\
\rowcolor{rowAlt} minimax-m2.5                   & 0.336 & 0.414 & 0.427 & 0.459 & 0.547 & 0.437 \\
\bottomrule
\end{tabular}
\arrayrulecolor{black}
\end{table}

\paragraph{Open-weight models.}
Several open-weight models land near the middle of the ranking: deepseek-v4-pro (\textsc{Avg} 0.494, C4 0.631) and mimo-v2.5-pro (\textsc{Avg} 0.490, C4 0.636) are competitive with the strongest Qwen3.6 and Gemma-4 rows.

\paragraph{Cost per condition (per user, USD, measured on the 42-user runs).}
The most expensive condition is Lifelong Memory (C4) with reasoning-heavy models: GPT-5 at Lifelong Memory costs $\sim$\$0.39 per user (\$16.4 per run). Gemini-3-Pro at Lifelong Memory costs $\sim$\$0.15 per user. Mid-size open models (Qwen3.6-27B, Gemma-4-31B) are roughly 7 to 35$\times$ cheaper: Qwen3.6-27B at Lifelong Memory is \$0.021 per user, Gemma-4-31B at Lifelong Memory is \$0.011 per user. The OpenAI, Vertex, and OpenRouter runs of the full ladder total $\approx$\$140 with caching enabled.

\paragraph{Excluded models.}
Two models (HY3-preview, Step-3.5-flash) were excluded after vendor-side instability during evaluation. Runs of the Gemini models through a second route are also excluded because they match the API route within $\pm 1.5$ points; we report only the API route to avoid double-counting. \texttt{gemini-3.1-pro-preview} reached the same C4 accuracy as \texttt{gemini-3-pro-preview} and is not listed separately.

\clearpage
\FloatBarrier
\section{Cross-Domain Transfer: Per-Model Diagonals}
\label{app:cdt}
This appendix collects the per-model diagonal breakdown for the Cross-Domain Transfer task. The main body keeps the S1 / S2 matrices (\cref{fig:domain-transfer}) and three findings; everything else is here.

\subsection{Per-model diagonals}
\label{sec:app-cdt-per-model}

\begin{figure}[!htbp]
\centering
\includegraphics[width=0.68\linewidth]{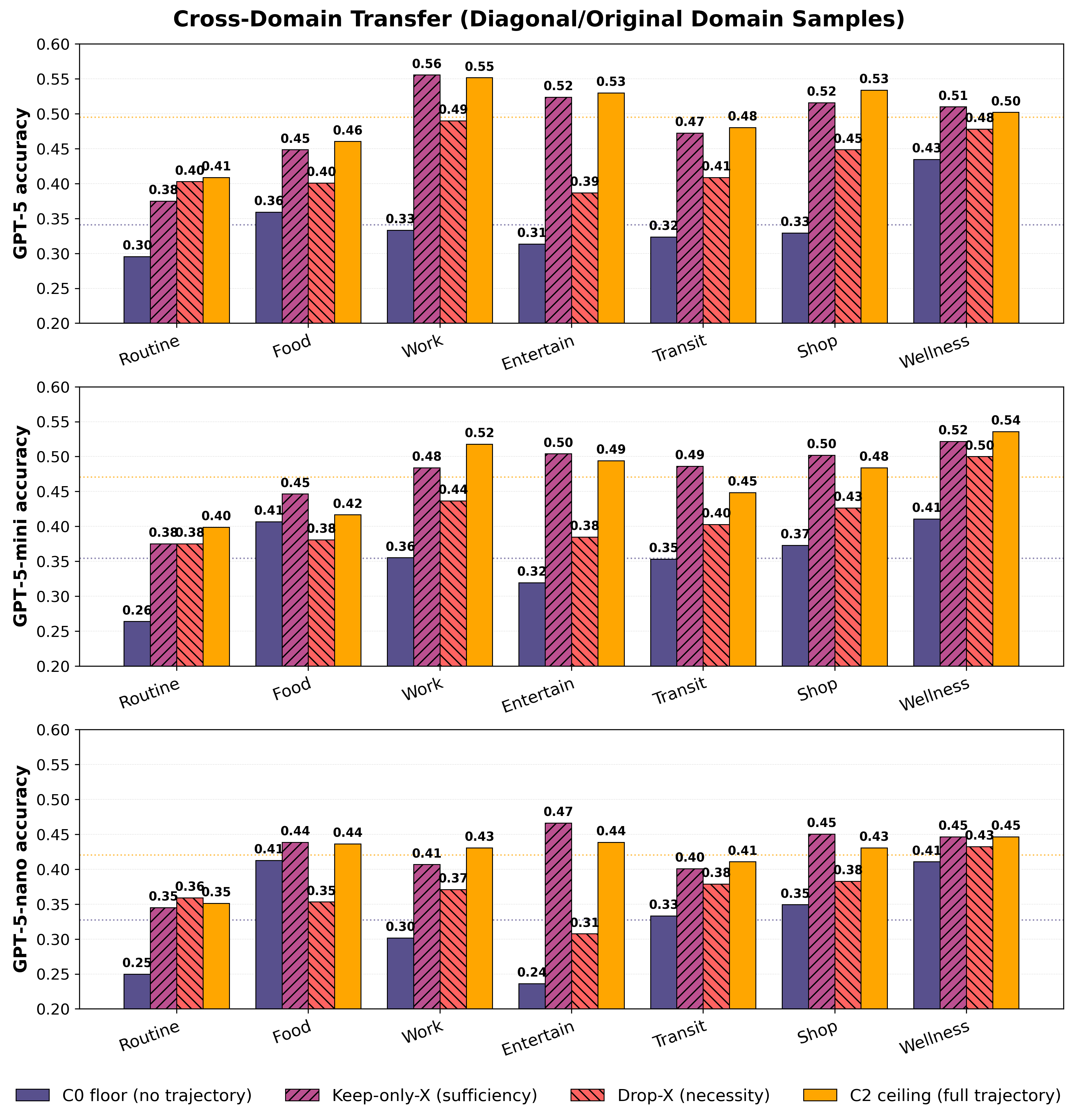}
\caption{\textbf{Cross-Domain Transfer per-model diagonals.} Absolute same-domain accuracy behind the diagonal of \cref{fig:domain-transfer} (C0 floor, keep-only, drop-one, C2 ceiling) for each GPT-5 size.}
\label{fig:cross-domain-per-model}
\end{figure}

\Cref{fig:cross-domain-per-model} shows, for each model, the absolute same-domain accuracies behind the diagonal of \cref{fig:domain-transfer}: the C0 floor, keep-only, drop-one, and the C2 ceiling.

\subsection{Daily Routine and Social Wellbeing are highly redundant}
\label{sec:app-cdt-redundant}

S1 drop-one $\Delta$ on \emph{Daily Routine} is only $-0.007$ on average and on \emph{Social Wellbeing} is $-0.025$. Removing them from Long Trajectory (C2) barely hurts because their signal (wake and bed times, social density) is already implicit in other lifestyle domains' events. The actionable consequence is that lightweight collection that skips fine-grained logging of these two domains would have minimal evaluation impact, complementing the headline finding that \emph{Leisure} is the dominant transfer source.

\clearpage
\FloatBarrier
\section{Validation Analyses}
\label{app:validation}
This appendix collects validation analyses of the synthesis pipeline (\cref{sec:app-val-anchor,sec:app-val-synth}), the Proactive Help contrasts (\cref{sec:app-val-significance}), the Cross-Domain Transfer finding (\cref{sec:app-val-cdt}), the RMR judge (\cref{sec:app-val-judge}), and the user selection for Proactive Help and Emotion-Aware Modeling (\cref{sec:app-val-selection}).

\subsection{Anchor-only versus full trajectory}
\label{sec:app-val-anchor}

To test whether the synthetic layer manufactures the preference signal, we rerun Implicit Preference Reasoning on an anchor-only trajectory that keeps every real anchor event and removes all synthetic events. \Cref{tab:anchor-only} compares it with the full trajectory at Long Trajectory (C2) and Lifelong Memory (C4). Without synthesis, C2 accuracy falls by 2 to 11 points, and the stronger models lose the most. The three stronger models then sit within 0.6 points of each other while Qwen3.5-9B stays last, so discrimination among strong models collapses. At C4 every model loses only 3 to 6 points and the model ordering is unchanged, because the survey insight and persona still carry strong preference signal. Synthesis therefore densifies behavioural evidence around real anchors rather than creating the signal, and the anchor-only trajectory is a much weaker evaluation surface at the current scale.

\begin{table}[h]
\centering
\caption{\textbf{Anchor-only versus full trajectory} (Implicit Preference Reasoning accuracy). The full-trajectory columns are taken from \cref{tab:matrix-b}.}
\label{tab:anchor-only}
\small
\begin{tabular}{@{}lcccccc@{}}
\toprule
Model & C2 anchor-only & C2 full & $\Delta$C2 & C4 anchor-only & C4 full & $\Delta$C4 \\
\midrule
Gemini-3-Pro     & 0.381 & 0.491 & $-0.110$ & 0.617 & 0.675 & $-0.058$ \\
Qwen3.6-27B      & 0.386 & 0.491 & $-0.105$ & 0.601 & 0.655 & $-0.054$ \\
Gemini-2.5-Flash & 0.387 & 0.447 & $-0.060$ & 0.539 & 0.580 & $-0.041$ \\
Qwen3.5-9B       & 0.308 & 0.328 & $-0.020$ & 0.469 & 0.500 & $-0.031$ \\
\bottomrule
\end{tabular}
\end{table}

\subsection{Sensitivity to the trajectory synthesizer}
\label{sec:app-val-synth}

Stages 1 to 3 of the synthesis pipeline use \texttt{gpt-5-mini} (\cref{app:dataset-construction}), which could favour models that share the generator's style. We compare Lifelong Memory (C4) accuracy in three settings: (a) the full 42-user set as released, (b) a random 15-user subset with the same synthesis, which controls for sample size, and (c) the same 15 users re-synthesized end to end with Qwen3.6-Plus as a drop-in replacement for \texttt{gpt-5-mini} in Stages 1 to 3. Between (b) and (c), per-model accuracy moves by at most 4.7 points, and the ranking Gemini-3-Pro $>$ Qwen3.6-27B $\approx$ Gemma-4-31B $>$ Claude-Sonnet-4.6 $>$ GPT-5 is identical in all three settings (\cref{tab:resynth}). Swapping the synthesizer therefore does not change the benchmark's model comparisons. The synthesizer's own family also gains no visible advantage, since GPT-5 and GPT-5-Mini rank in the lower half of the 54 models (\cref{tab:matrix-a}).

\begin{table}[h]
\centering
\caption{\textbf{Implicit Preference Reasoning C4 accuracy under three synthesis settings.} Column (a) is taken from \cref{tab:matrix-a,tab:matrix-b}.}
\label{tab:resynth}
\small
\begin{tabular}{@{}lccc@{}}
\toprule
Model & \makecell{(a) 42 users\\GPT synthesis} & \makecell{(b) 15 users\\GPT synthesis} & \makecell{(c) 15 users\\Qwen3.6-Plus synthesis} \\
\midrule
Gemini-3-Pro      & 0.675 & 0.688 & 0.641 \\
Qwen3.6-27B       & 0.655 & 0.640 & 0.622 \\
Gemma-4-31B       & 0.652 & 0.636 & 0.618 \\
Claude-Sonnet-4.6 & 0.631 & 0.630 & 0.598 \\
GPT-5           & 0.585 & 0.595 & 0.551 \\
\bottomrule
\end{tabular}
\end{table}

\subsection{Statistical significance on Proactive Help}
\label{sec:app-val-significance}

To check that the image-versus-text contrasts in \cref{sec:results-proactive} are not sampling noise, we repeat the Proactive Help evaluation three times for Qwen3.6-27B, Gemma-4-31B, and GPT-5 and test the per-user F1 difference with an exact paired permutation test at the user level over the 15 covered users (\cref{tab:proactive-sig}). All three contrasts are significant at $p < 0.01$, and their signs differ by family, with image input hurting Qwen and Gemma and helping GPT-5.

\begin{table}[h]
\centering
\caption{\textbf{Proactive Help, image versus text, over three runs.} U-F1 is the per-user mean over the 15 covered users, reported as mean $\pm$ standard deviation over the three runs; $p$ is from an exact paired permutation test at the user level.}
\label{tab:proactive-sig}
\small
\begin{tabular}{@{}lcccc@{}}
\toprule
Model & U-F1 text & U-F1 image & $\Delta$U-F1 (image $-$ text) & $p$ (paired) \\
\midrule
Qwen3.6-27B & $42.7 \pm 1.7$ & $25.2 \pm 1.6$ & $-17.6$ & $< 0.001$ \\
Gemma-4-31B & $40.6 \pm 1.3$ & $36.3 \pm 0.9$ & $-4.3$  & $0.003$ \\
GPT-5     & $39.3 \pm 1.7$ & $53.2 \pm 1.6$ & $+13.8$ & $< 0.001$ \\
\bottomrule
\end{tabular}
\end{table}

\subsection{Cross-Domain Transfer beyond GPT-5}
\label{sec:app-val-cdt}

We extend the S1 drop-one diagonal to Qwen3.6-27B and Gemma-4-31B (\cref{tab:cdt-families}). In all three families, removing \texttt{leisure} from the Long Trajectory costs the most same-domain accuracy (10.5 to 15.5 points), followed by \texttt{work\_study}, while \texttt{daily\_routine} and \texttt{social\_wellbeing} remain nearly redundant. The top two and the bottom two domains are the same in all three families, so the leisure finding is a property of the benchmark rather than of one model family.

\begin{table}[h]
\centering
\caption{\textbf{Cross-Domain Transfer S1 (drop-one) diagonal across three model families.} Each entry is the same-domain accuracy change when the named domain is removed from Long Trajectory (C2). The GPT-5 column averages the three GPT-5 sizes (\cref{fig:domain-transfer}).}
\label{tab:cdt-families}
\small
\begin{tabular}{@{}lccc@{}}
\toprule
Domain dropped from C2 & GPT-5 & Qwen3.6-27B & Gemma-4-31B \\
\midrule
\texttt{daily\_routine}    & $-0.007$ & $-0.018$ & $-0.004$ \\
\texttt{food\_dining}      & $-0.060$ & $-0.062$ & $-0.035$ \\
\texttt{work\_study}       & $-0.068$ & $-0.108$ & $-0.075$ \\
\texttt{leisure}           & $\mathbf{-0.128}$ & $\mathbf{-0.155}$ & $\mathbf{-0.105}$ \\
\texttt{commuting\_travel} & $-0.050$ & $-0.088$ & $-0.058$ \\
\texttt{shopping}          & $-0.064$ & $-0.069$ & $-0.041$ \\
\texttt{social\_wellbeing} & $-0.025$ & $-0.039$ & $-0.016$ \\
\bottomrule
\end{tabular}
\end{table}

\subsection{Human validation of the RMR judge}
\label{sec:app-val-judge}

The rationale match rate relies on an LLM judge (GPT-5). Five annotators independently scored 48 Proactive Help samples, each consisting of the proactive context, the participant's own reason, and a model-predicted reason. We compare the judge with each annotator and with the human majority vote (\cref{tab:rmr-judge}). Agreement among humans is almost perfect (mean pairwise Cohen's $\kappa = 0.855$, Fleiss' $\kappa = 0.853$), and the judge agrees with the majority vote at $\kappa = 0.909$ and 95.8\% accuracy, inside the human agreement band on the scale of \citet{landis1977measurement}. The ICC(2,1) values fall in the good range of \citet{koo2016guideline}.

\begin{table}[h]
\centering
\caption{\textbf{RMR judge validation on 48 Proactive Help samples.} The LLM judge sits inside the human agreement band.}
\label{tab:rmr-judge}
\small
\begin{tabular}{@{}llc@{}}
\toprule
Comparison & Metric & Value \\
\midrule
Human vs.\ human (mean pairwise)  & Cohen's $\kappa$ & 0.855 \\
Human vs.\ human (5 raters)       & Fleiss' $\kappa$ & 0.853 \\
Human vs.\ human (5 raters)       & ICC(2,1)         & 0.856 \\
Judge vs.\ human majority vote    & Cohen's $\kappa$ & 0.909 \\
Judge vs.\ human majority vote    & Accuracy         & 0.958 \\
Judge vs.\ human majority vote    & Pearson $r$      & 0.913 \\
5 humans and the judge (6 raters) & ICC(2,1)         & 0.865 \\
\bottomrule
\end{tabular}
\end{table}

\subsection{User selection for Proactive Help and Emotion-Aware Modeling}
\label{sec:app-val-selection}

Both tasks are scored only on participant-flagged real anchors, so a participant enters a task pool only after contributing enough annotated multimodal anchors. For Proactive Help, a user qualifies if at least one logged day contains an anchor with both a photograph and a non-empty \texttt{agentHelpReason}. For Emotion-Aware Modeling, a user qualifies if, within their last five logged days, at least one anchor has both a photograph and non-empty \texttt{moodKeywords}. Each rule admits 15 of the 42 users, who contribute 254 Proactive Help anchors (111 positive, 143 negative) and 123 emotion samples (78 positive, 22 negative, 23 mixed), and eight users belong to both pools. More anchors exist in principle, but we keep these stricter multimodal rules (the recent window applies to the emotion task) for coherent evaluation, and the pools will grow with the next recruitment round.

To test whether the rule selects users whose preferences are easier or harder to infer, we compare the Implicit Preference accuracy of the covered users with that of the remaining 27 users, using the per-user outputs of the 26 Bedrock-served models (\cref{tab:selection-bias}). The covered users score 1.0 to 1.3 points higher at both C2 and C4, and a user-level permutation test over 5{,}000 random 15-user subsets finds no significant difference ($p \geq 0.5$ in all four cases). We therefore find no evidence that the selection rule picks systematically easier or harder users on the shared preference task.

\begin{table}[h]
\centering
\caption{\textbf{Selection-bias check.} Implicit Preference accuracy (\%) of the 15 users covered by each task versus the other 27 users, averaged over the 26 Bedrock-served models. $p$ is from a user-level permutation test over 5{,}000 random 15-user subsets.}
\label{tab:selection-bias}
\small
\begin{tabular}{@{}llcccc@{}}
\toprule
Task pool & Condition & Covered (15) & Others (27) & Gap & $p$ \\
\midrule
Proactive Help & Long Trajectory (C2) & 49.5 & 48.5 & $+1.0$ & 0.62 \\
Proactive Help & Lifelong Memory (C4) & 64.7 & 63.4 & $+1.3$ & 0.53 \\
Emotion-Aware  & Long Trajectory (C2) & 49.7 & 48.4 & $+1.3$ & 0.54 \\
Emotion-Aware  & Lifelong Memory (C4) & 64.6 & 63.4 & $+1.2$ & 0.57 \\
\bottomrule
\end{tabular}
\end{table}

\clearpage
\FloatBarrier
\section{Prompts, Schemas, and Reproducibility}
\label{app:prompts}
\subsection{Stage prompts}

All four stages of the synthesis pipeline use a single dedicated system prompt per stage; the full prompt library is released alongside the benchmark. We summarise the load-bearing constraints below.

\paragraph{Stage 0, image grounding.}
The vision model is told that each photograph is drawn from a participant's daily log; it must emit a one-sentence third-person caption of what the participant is doing, together with a more abstracted caption that is rich enough to ground a synthetic event but robust to identifying detail. The prompt explicitly forbids identifying persons by name or appearance and forbids transcribing private text visible in the image.

\paragraph{Stage 1, persona and survey insight.}
The reasoning model receives the participant's structured persona, all 100 survey responses, a locally computed digest of behavioural traits (chronotype, sleep-after-midnight rate, and similar), and a sample of image captions. It produces an enhanced profile across eighteen axes, a uniqueness signature, a six-to-twelve-sentence persona narrative, and a survey-insight summary that contains implicit-preference paragraphs, a list of cross-axis deep inferences, and a domain-indexed view. The prompt forbids the survey-insight summary from naming any question text or option letter, so that downstream stages cannot regurgitate the survey verbatim.

\paragraph{Stage 2, day skeleton and clue plan.}
For each fourteen-day window the model produces a day-level skeleton (per-day theme, narrative arc, dominant mood, energy curve, and 10 to 18 planned events) together with a planting plan that assigns each preference question to one or more events through which behavioural evidence should appear. The prompt enforces three invariants: every real anchor must be retained; synthetic events must be at least as many as anchors unless the day already exceeds fifteen events; and clues must be planted across at least two distinct days when feasible.

\paragraph{Stage 3, trajectory expansion.}
The model expands the skeleton in five-day chunks. For anchor events, only the third-person narrative paraphrase is authored; the original log content is reproduced verbatim. For synthetic events, the model has full content freedom but is conditioned on the day's theme, narrative arc, the participant's uniqueness signature, and any planting plan entries that target the event. The prompt enumerates the seven planted-clue kinds (timing detail, consequence, decision frame, post-event mood, social context, app choice, and tiny detail) and explicitly forbids restating survey questions or option letters.

\FloatBarrier
\section{Inference Protocol}
\label{app:inference}
All 54 evaluated models are queried through four providers: the OpenAI API, Google Vertex AI or AI Studio, OpenRouter, and Amazon Bedrock. The released router exposes a single chat-completion interface that normalises message formatting, multimodal content, and response parsing across providers, and the Bedrock runs use the same prompt builder and parser, ensuring that every model receives identical prompts, identical system messages, and identical image payloads.

We use three inference settings for every model and every condition. (i) Temperature is set to 0. Models that reject a non-default temperature, such as the GPT-5 family, and all Bedrock-served models (no temperature is sent) run at the provider default, so their results are single samples rather than deterministic decodes. (ii) Reasoning follows a per-model setting. Gemini-2.5-Flash and Gemini-3-Flash use a thinking budget of 0, DeepSeek-V4-Pro, Kimi-K2.6, GLM-5, and MiniMax-M2.5 / M2.7 receive an OpenRouter reasoning cap of 10{,}000 tokens, Bedrock-served models use the low reasoning-effort tier where it is available (Claude-Opus-4.5 runs at its default), and the remaining models, including the GPT-5 family, use their provider's default. (iii) The output token budget is 16{,}000 tokens, raised to 32{,}000 or 48{,}000 for the models with a reasoning cap, which is large enough to fit the full structured response on every task without truncation.

All other generation parameters (top-p, frequency penalty, presence penalty) are left at provider defaults. Image inputs are passed through the router as base64-encoded payloads at a single resolution tier across providers. A response cache keyed on the model identifier, the prompt content, and the generation parameters guarantees that re-runs reproduce the exact response, and the cache is invalidated whenever any cap or parameter changes.

\paragraph{Structured JSON output.}
Every model on every task is asked to return a structured JSON object rather than free-form text. The router requests provider-native JSON mode (response-format JSON on OpenAI-compatible routes, response MIME type application/json for Gemini) and drops it for models that reject it, and the Bedrock runs do not use JSON mode. Each task ships a fixed JSON schema describing the expected fields (an answer letter for the Implicit Preference task, a binary trigger flag with a short rationale for the Proactive Help task, a polarity label and a multi-label mood vector for the Emotion-Aware task, and a per-question answer object for the Cross-Domain Transfer ablations). When a response is not valid JSON, the parser falls back to the first brace-delimited block and then to a regular expression over question-letter pairs, and a question that none of these steps can read is scored as wrong. JSON output also limits the ambiguity of regex-based answer extraction, which is a known confound in chat-completion benchmarking.

Two practical consequences follow. First, every number obtained through the released harness is reproducible from its configuration file and response cache. Second, no model is tuned, every headline number comes from a single run (the only repeated sampling is the three-run check in \cref{sec:app-val-significance}), and reasoning follows the per-model settings above; differences in reported accuracy and F1 reflect differences in model capability under this evaluation contract.

\FloatBarrier
\section{Extended Discussion}
\label{app:extended-discussion}

\paragraph{What \textsc{Jarvis-Bench} reveals.}
The user-memory ladder isolates three distinct contributors to personal-agent performance: dense behavioural context (the largest single-step gain, No Memory to Long Trajectory $\sim$12 points), explicit persona text (modest, Long Trajectory to Persona Memory $\sim$3 to 4 points), and a distilled survey-insight oracle that approximates an ideally-prepared profile (large, Persona Memory to Lifelong Memory $\sim$10 to 13 points). The dominant lift from Long Trajectory (full lived trajectory, no persona text) shows that today's frontier LLMs can read a person from their behaviour, but only up to the 0.732 Lifelong-Memory ceiling held by GPT-6-Astra. The Cross-Domain Transfer ablation localises this signal further: \emph{Leisure} carries disproportionately rich preference information that transfers across other lifestyle domains, while \emph{Daily Routine} and \emph{Social Wellbeing} are largely redundant.

\paragraph{What overall accuracy hides.}
Implicit Preference leadership belongs to GPT-6-Astra and GPT-6.1-Sol; emotion-text leadership flips to GPT-5; emotion-image flips again to Qwen3.6-27B; proactive pool F1 is led by Qwen3.6-35B-A3B (text) and GPT-5-Mini (image); pooled rationale match rate on the proactive task is led by GPT-5 (text) and GPT-5-Mini (image). \emph{No single family wins all four task surfaces.} Multimodality helps the emotion task (image $\geq$ text on nearly every model) but hurts the proactive task for most Qwen models (Qwen3.5-9B loses 43 pool-F1 points and Qwen3.6-27B 19 per-user points), a divergence we trace in \cref{sec:results-proactive} to the difference between scene-judgable affect and timing-conditioned trigger judgement. \textsc{Jarvis-Bench} is therefore most informative when its task surfaces are read together: a model that excels on the survey task may still be a poor proactive companion.

\paragraph{Implications for personal-agent training.}
Three implications follow. First, training and evaluation for systems that intend to act proactively cannot be limited to the survey task; the proactive task surface, with its grounded participant-confirmed help ground truth, is essential. Second, the gap between binary-prediction F1 and rationale match rate on the proactive task suggests that explanation quality, not just decision quality, deserves first-class evaluation. Third, the Cross-Domain Transfer ablation suggests that production cold-start for personal agents can rely on the \emph{Leisure} lifestyle domain (entertainment events in the trajectory) as a high-leverage starting point.

\paragraph{Contribution as evaluation methodology.}
We position the contribution as evaluation methodology grounded in real users. The dataset is a substrate; the reusable artefacts are (a) the user-memory ladder (No Memory $\to$ Short Trajectory $\to$ Long Trajectory $\to$ Persona Memory $\to$ Lifelong Memory), which decomposes any personalization claim into trajectory-density, persona, and oracle-context contributions; (b) the S1 / S2 ablation, which converts cross-domain transfer from a hypothesis into a direct measurement applicable to any multi-domain trajectory benchmark; (c) the per-user mean vs.\ pool dual-reporting protocol, which exposes the anchor-rich vs.\ anchor-sparse distortion that single-pool numbers hide on the proactive task; (d) the longitudinal participant-feedback loop (\cref{sec:app-loop}), which converts a self-tracking corpus into a dataset whose qualified users have approved their own released trajectories.

\paragraph{Ethics, consent, and release.}
All participants provided informed consent under an IRB-approved protocol. Calendar dates are stripped, locations and surnames are removed, photographs are described in the trajectory text only by third-person captions, and a privacy blacklist is enforced by an automatic validator. We release the benchmark under a dataset licence (CC BY-NC 4.0) that restricts use to non-commercial research.



\end{document}